\PassOptionsToPackage{table,svgnames}{xcolor}
\documentclass[11pt]{article}

\usepackage{acl}

\usepackage{times}
\usepackage{latexsym}
\usepackage[T1]{fontenc}
\usepackage[utf8]{inputenc}
\usepackage{microtype}
\usepackage{inconsolata}
\usepackage{graphicx}
\usepackage{amsmath}
\usepackage{booktabs}
\usepackage{cuted}
\usepackage{multirow}
\usepackage{enumitem}
\usepackage{xcolor}
\usepackage{amssymb}
\usepackage{xspace}
\usepackage{listings}
\usepackage{fancyhdr}
\usepackage{wrapfig}
\usepackage{pifont}
\usepackage[inkscapelatex=false]{svg}
\usepackage{subcaption}
\usepackage{array}
\usepackage{tabularx}
\usepackage{makecell}
\usepackage{colortbl}
\usepackage{stfloats}


\svgsetup{inkscapearea=drawing}
\definecolor{darkblue}{rgb}{0, 0, 0.5}
\hypersetup{colorlinks=true, citecolor=darkblue, linkcolor=darkblue, urlcolor=darkblue}
\newcommand{\bench}{TempCloze\xspace}

\title{TempCloze: Can Video-LLMs Identify the Missing Middle?}

\author{
  \textbf{Wenqi Pei\textsuperscript{1,*}} \quad
  \textbf{Henry Hengyuan Zhao\textsuperscript{2,*}} \quad
  \textbf{Yilai Liu\textsuperscript{1,*}} \quad
  \textbf{Jiahao Meng\textsuperscript{3}} \\
  \textbf{Han Chen\textsuperscript{2}} \quad
  \textbf{Ziyu Wang\textsuperscript{3}} \quad
  \textbf{Hongyang Du\textsuperscript{1,\(\dagger\)}} \\
  \\
  \textsuperscript{1}The University of Hong Kong \quad
  \textsuperscript{2}National University of Singapore \quad
  \textsuperscript{3}Peking University
}

\begin{document}
\maketitle
\begingroup
\renewcommand{\thefootnote}{}
\makeatletter
\def\Hy@footnote@currentHref{Hfootnote.authornote}
\makeatother
\footnotetext{\mbox{\textsuperscript{*} Equal contribution.\enspace\textsuperscript{\(\dagger\)} Corresponding author.}}
\endgroup
\begin{abstract}
Temporal reasoning benchmarks for Video-LLMs are often mediated by language, leaving room for linguistic shortcuts from option wording, answer correlations, or language priors. To reduce such shortcuts, we introduce \textbf{\bench}\footnote{\url{https://github.com/CedricPei/Temporal-Cloze}}, a video cloze benchmark for evaluating visual temporal reasoning in Video-LLMs. Given the beginning and ending clips of a video, models must identify the true missing middle from four candidates. \bench contains 1,521 carefully filtered videos from seven sources, mainly long-take and egocentric videos. We construct same-source distractors along three dimensions: \textit{Semantic} asks what event should happen, \textit{Alignment} probes when it should occur, and \textit{Progression} tests how it should unfold, while shared scenes and objects reduce appearance cues. Our evaluation of 10 proprietary and 21 open-source Video-LLMs reveals Alignment as the primary bottleneck: models often recognize plausible semantic content and local event progression but struggle with temporal alignment. We further conduct error pattern and behavioral sensitivity analyses on TempCloze-Mixed and TempCloze-Hard with four representative models to examine where errors arise and how candidate order, context direction, visible span, frame density, and test-time scaling influence model choices.
\end{abstract}

\begin{figure*}[t]
  \centering
  \includegraphics[width=0.95\textwidth]{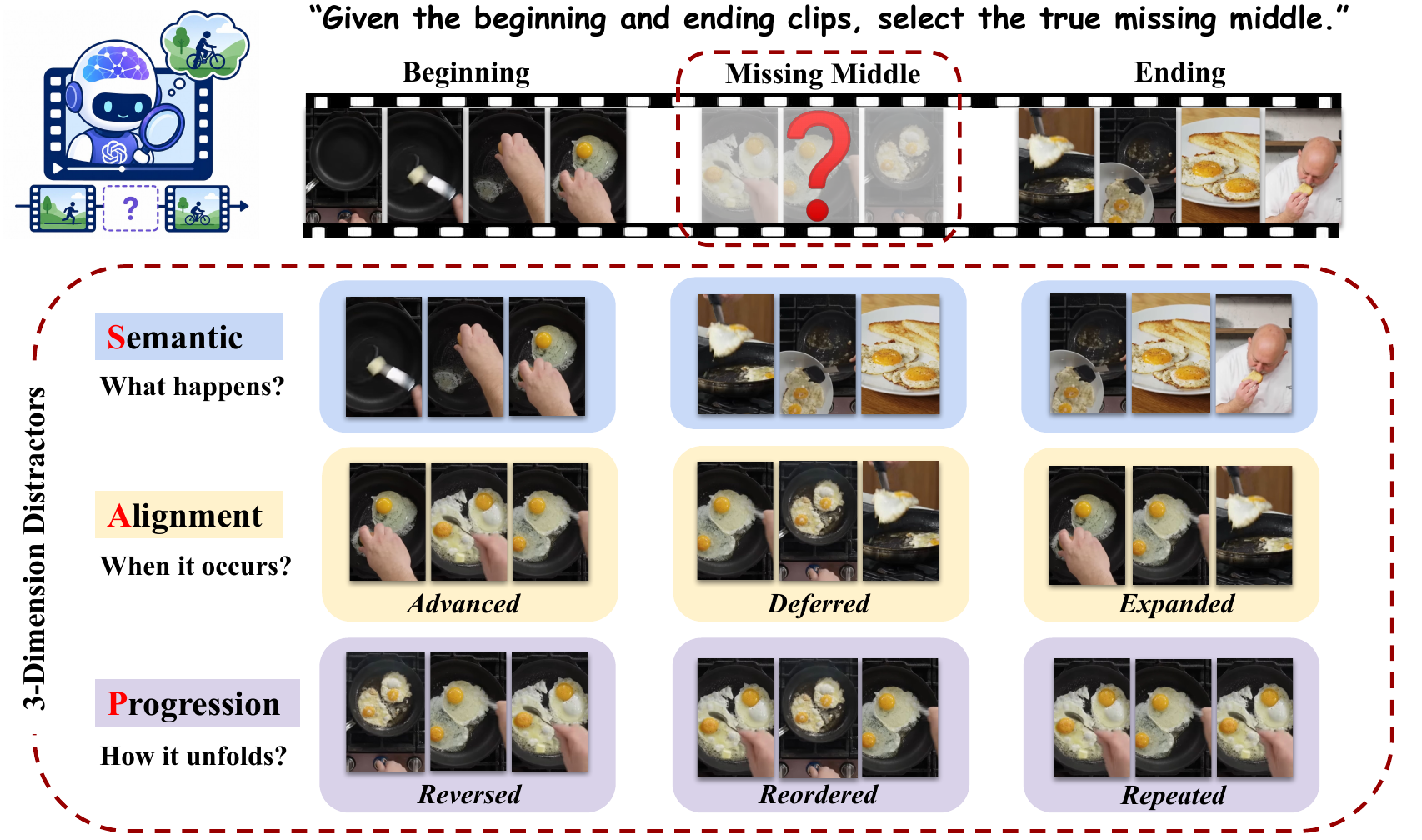}
  \vspace{-0.6em}
  \caption{Overview of \bench. Given the beginning and ending clips, models are asked to identify the true missing middle. Distractors are constructed along three dimensions: Semantic, Alignment and Progression.}
  \label{fig:main}
  \vspace{-0.8em}
\end{figure*}

\vspace{-1em}
\section{Introduction}

Video Large Language Models (Video-LLMs) have become increasingly capable of video temporal reasoning with advances in multimodal architectures and model scaling~\citep{li2023blip2,zhao2025interfeedback,meng2025open, meng2026watch}. Though Video-LLMs have achieved strong performance on temporal reasoning benchmarks like TempCompass and TVBench~\citep{liu2024tempcompass,cores2024tvbench}, the benchmarks are moving beyond conventional VideoQA to cover longer contexts and more fine-grained questions~\citep{fu2025videomme,zhang2025universal,meng2026videozerobench,xu2026towards}.

However, these evaluations are still mediated by language. Models need to choose among textual options or produce detailed captions~\citep{lei2018tvqa, maaz2024videochatgpt}. Such formats can introduce linguistic shortcuts, allowing models to exploit option wording, answer correlations, or language priors to obtain competitive scores~\citep{xiao2024vgqa, balepur2024artifacts}. For example, pretrained language models can exceed the random baseline by more than 25\% without multimodal context~\citep{yang2020what}. Accordingly, a more direct evaluation should ask models to compare visual evidence in video segments rather than textual descriptions. This motivates a benchmark for the evaluation of visual temporal reasoning in Video-LLMs.

To address the gap, we introduce \bench, a video cloze benchmark centered on the question: \emph{Can Video-LLMs Identify the Missing Middle?} Given the beginning and ending clips of a video, models are required to identify the true missing middle from a set of candidates. The correct choice is not merely visually related to the video, but occupies the proper temporal position between the observed clips. Identifying the middle requires models to infer what happens, when it occurs, and how it unfolds. This setup encourages models to reason over time from what they see, reducing the influence of linguistic shortcuts in evaluation.

To evaluate distinct aspects of visual temporal reasoning, \bench constructs multi-dimension distractors from the same source. They share scenes and objects with the original video. This reduces appearance cues and encourages temporal comparison rather than visual matching. Moreover, distractors cover three dimensions: \textbf{Semantic (S)} tests what event should happen by contrasting the correct event with other non-overlapping intervals, \textbf{Alignment (A)} probes when the event should occur by shifting or expanding the ground-truth span, and \textbf{Progression (P)} evaluates how the event should develop by reversing, reordering, or repeating the segment. Collectively, we assess event semantics, temporal alignment, and temporal progression.

We construct \bench from seven sources covering long-take~\citep{xiong2024lvd2m}, egocentric~\citep{yang2025egolife}, and fine-grained motion~\citep{tu2025favorbench} videos. After filtering out overlength and low-quality videos, we employ a reasoning LLM to examine their captions and exclude those unsuitable for cloze formatting. We then use optical flow to remove largely static videos~\citep{farneback2003two}. The resulting benchmark contains 1,521 videos, each instantiated along three dimensions. We evaluate 10 proprietary and 21 open-source Video-LLMs and observe a consistent bottleneck: while models can identify plausible event content and recognize its progression, they struggle substantially with temporal alignment. Additionally, we construct two auxiliary subsets: TempCloze-Mixed, which mixes distractors across temporal dimensions, and TempCloze-Hard, which consists of the most widely misclassified instances.

We further conduct Error Pattern Analysis and Behavioral Sensitivity Analysis to examine where errors arise and what factors affect model choices. Dimension-specific errors show that Alignment failures are dominated by Expanded, while Progression failures are marked by Reversed. Mixed-dimension errors show that Alignment is the most misleading alternative. We also find that selections are unstable under candidate reordering but with relatively stable performance. Furthermore, models rely more on beginning context than ending. They depend strongly on endpoint information, and adding frame density can dilute the information. Test-time scaling brings model-dependent gains, but does not change the relative ordering among dimensions. These analyses offer broader insights into the performance and limitations of visual temporal reasoning in existing Video-LLMs.

Our contributions are summarized as follows:
\vspace{-0.35em}
\begingroup
\setlength{\leftmargini}{6pt}
\begin{itemize}
    \setlength{\itemsep}{0pt}
    \item We introduce \bench, a video cloze benchmark comprising 1,521 videos to evaluate visual temporal reasoning in Video-LLMs. Each video is instantiated with three dimensions: Semantic, Alignment, and Progression.
    \item We evaluate 10 proprietary and 21 open-source Video-LLMs and identify the Alignment dimension as the main bottleneck.
    \item We conduct Error Pattern Analysis and Behavioral Sensitivity Analysis, revealing where errors arise and how some factors influence model choices.
\end{itemize}
\endgroup

\section{Related Work}

\subsection{Temporal Reasoning Benchmarks}

Temporal reasoning benchmarks for video have grown increasingly complex in recent years. One line comes from general video understanding benchmarks that include temporal reasoning as part of a broader evaluation. Early VideoQA datasets such as TGIF-QA~\citep{jang2017tgifqa}, NExT-QA~\citep{xiao2021nextqa} and ActivityNet-QA~\citep{yu2019activitynetqa} established this setting with questions about explicit actions and spatio-temporal relations. Recent long-context benchmarks such as LongVideoBench~\citep{wu2024longvideobench}, MVBench~\citep{li2024mvbench} and VideoMME~\citep{fu2025videomme} broaden coverage across diverse tasks, requiring evidence aggregation over extended dynamics rather than isolated cues. Another line focuses more directly on temporal structure, with TemporalBench~\citep{cai2024temporalbench}, TVBench~\citep{cores2024tvbench}, and TempCompass~\citep{liu2024tempcompass} testing event ordering, moment localization, and fine-grained motion. However, these benchmarks are still language-mediated. This leaves room for linguistic shortcuts, where models can benefit from surface regularities~\citep{ma2024robustvisualquestionanswering,zhong2022videoquestionansweringdatasets}. \bench asks models to identify the missing middle from video candidates, shifting evaluation toward visual temporal reasoning.

\begin{table}[bt]
\vspace{-0.5em}
\centering
\caption{Composition over 1,521 videos. Left: videos per source; Right: duration distribution in seconds.}
\label{tab:sources}
\vspace{0.15em}
\makebox[\columnwidth][c]{%
\begin{minipage}[c][1.30in][c]{0.285\columnwidth}
\raggedleft
\raisebox{0.06in}[0pt][0pt]{%
\resizebox*{!}{1.06in}{%
\begingroup
\renewcommand{\arraystretch}{0.95}
\begin{tabular}{@{}l r@{}}
\toprule
\textbf{Source} & \textbf{\#} \\
\midrule
LVD-2M & 515 \\
EgoLife & 437 \\
MiraData & 198 \\
FAVOR & 145 \\
CaReBench & 94 \\
Video-TT & 89 \\
Daily-Omni & 43 \\
\bottomrule
\end{tabular}
\endgroup
}%
}%
\end{minipage}\hspace{0.003\columnwidth}%
\begin{minipage}[c][1.30in][c]{0.700\columnwidth}
\raggedright
\includegraphics[width=\linewidth,height=1.30in,keepaspectratio]{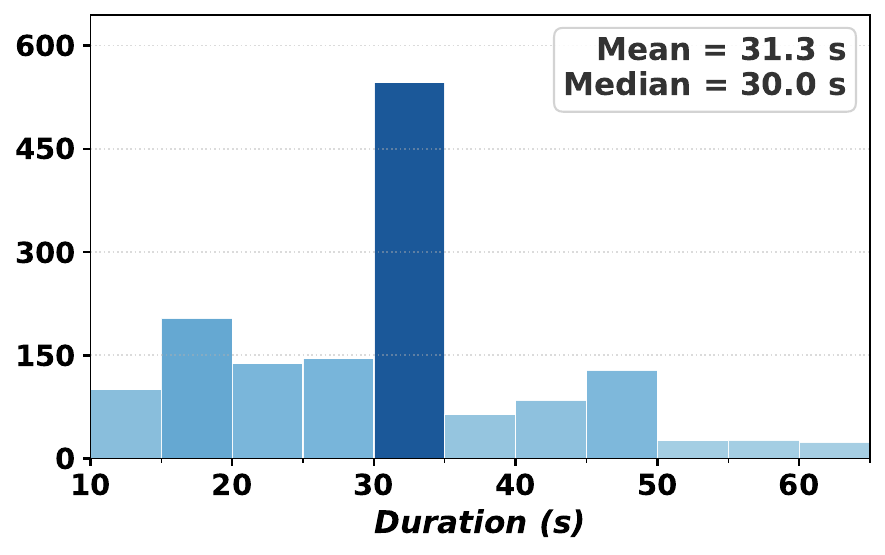}%
\end{minipage}%
}
\vspace{-2em}
\end{table}

\subsection{Cloze for Video Understanding} 
Cloze tasks remove part of a context and infer the missing content~\citep{taylor1953cloze}. MovieFIB~\citep{maharaj2017moviefib} asks models to fill blanks in movie descriptions and FIBER~\citep{castro2022fiber} extends this with constituent-level blanks and multiple valid answers. While they evaluate video understanding through textual completion, cloze-style objectives have also been used for video representation learning. VideoBERT~\citep{sun2019videobert} is an early pre-training method that applies masked prediction to learn temporal correspondences between video and text. VCP~\citep{luo2020vcp} learns by predicting transformations applied to withheld video clips. MaskFeat~\citep{wei2022maskfeat} masks parts of the video input and predicts features of the regions, while VideoMAE~\citep{tong2022videomae} reconstructs masked video patches for pre-training. These works demonstrate the value of cloze formatting, but they use it as a training objective. \bench instead formulates video cloze for the evaluation of temporal reasoning in Video-LLMs.

\newcommand{\cmark}{\ding{51}}
\newcommand{\xmark}{\ding{55}}

\definecolor{rankone}{HTML}{F6D4B5}
\definecolor{ranktwo}{HTML}{D7E8F5}
\definecolor{rankthree}{HTML}{E1ECD8}
\newcommand{\Best}[1]{\cellcolor{rankone}\textbf{#1}}
\newcommand{\Second}[1]{\cellcolor{ranktwo}\underline{#1}}
\newcommand{\Third}[1]{\cellcolor{rankthree}#1}
\newcommand{\Baseline}[1]{\textbf{#1}}
\newcommand{\RankLegend}[2]{\begingroup\setlength{\fboxsep}{1pt}\colorbox{#1}{\strut #2}\endgroup}

\newcommand{\BlockRow}[1]{
    \specialrule{0.55pt}{0.2em}{0pt}
    \rowcolor{gray!16}[0pt][0pt]
    \multicolumn{9}{@{}c@{}}{\rule{0pt}{2.45ex}\normalsize\itshape #1}\\
    \specialrule{0.55pt}{0pt}{0.2em}
}

\newcommand{\BlockBlue}[1]{
    \specialrule{0.55pt}{0.2em}{0pt}
    \rowcolor{blue!15}[0pt][0pt]
    \multicolumn{9}{@{}c@{}}{\rule{0pt}{2.45ex}\normalsize\itshape #1}\\
    \specialrule{0.55pt}{0pt}{0.2em}
}

\newcolumntype{Y}{>{\centering\arraybackslash}X}

\begin{table*}[t]
\centering
\caption{
Main results on \bench. Dimension Accuracy reports Semantic, Alignment, and Progression accuracy; Cumulative Accuracy reports the fraction of videos with at least one, at least two, or all three dimensions correct.
\RankLegend{rankone}{Peach}, \RankLegend{ranktwo}{blue}, and \RankLegend{rankthree}{sage} mark the top three; \textbf{Bold} and \underline{underlined} values mark the best and second-best.
Human\textsuperscript{\(\star\)} Baseline is measured on 100 randomly selected videos, and the detailed protocol is provided in Appendix~\ref{sec:human_baseline}.
}
\label{tab:main_results}
\footnotesize
\renewcommand{\arraystretch}{1}
\setlength{\tabcolsep}{2.8pt}

\begin{tabularx}{\textwidth}{@{} l @{\hspace{3pt}} c *{7}{Y} @{}}
\toprule
\multicolumn{2}{c}{\textbf{Model}} &
\multicolumn{4}{c}{\textbf{Dimension Accuracy (\%)}} &
\multicolumn{3}{c}{\textbf{Cumulative Accuracy (\%)}} \\
\cmidrule(lr){1-2}\cmidrule(lr){3-6}\cmidrule(lr){7-9}
Name & Think & S$\uparrow$ & A$\uparrow$ & P$\uparrow$ & Mean$\uparrow$ & $\geq\!1\uparrow$ & $\geq\!2\uparrow$ & $3/3\uparrow$ \\

\BlockRow{Proprietary Models}
Seed1.8 & \cmark & \Second{95.07} & \Best{76.92} & \Best{93.75} & \Best{88.58} & \Best{99.54} & \Best{95.40} & \Best{70.81} \\
Qwen3.5-Plus & \cmark & \Third{90.72} & \Second{76.32} & \Third{90.20} & \Second{85.74} & \Third{98.75} & \Third{91.25} & \Second{67.22} \\
Seed1.8 & \xmark & \Best{96.25} & 61.93 & \Second{92.11} & \Third{83.43} & \Second{99.35} & \Second{94.02} & \Third{56.94} \\
Gemini2.5-Pro & \cmark & 90.33 & \Third{66.67} & 67.78 & 74.92 & 97.44 & 83.49 & 43.95 \\
Gemini2.5-Flash & \xmark & 82.18 & 59.83 & 74.82 & 72.28 & 94.61 & 78.17 & 44.05 \\
GPT5.4 & \xmark & 68.11 & 37.41 & 65.15 & 56.89 & 89.35 & 60.16 & 21.17 \\
Claude4.6-Sonnet & \xmark & 55.69 & 28.80 & 70.94 & 51.81 & 86.12 & 52.66 & 16.63 \\
Claude4.6-Opus & \cmark & 49.47 & 39.34 & 62.89 & 50.57 & 81.97 & 50.85 & 18.88 \\
Gemini3-Flash & \xmark & 61.08 & 40.37 & 48.13 & 49.86 & 84.09 & 49.57 & 15.91 \\
Seed1.6 & \xmark & 64.59 & 17.83 & 55.40 & 45.93 & 81.65 & 46.68 & 9.38 \\
Grok4.1 & \xmark & 24.52 & 24.06 & 23.73 & 24.11 & 57.53 & 14.07 & 0.72 \\
\textbf{Avg.} & & 70.73 & 48.13 & 67.72 & 62.19 & 88.22 & 65.12 & 33.24 \\
\addlinespace[2pt]

\BlockRow{Open-source Models}
Qwen3.5-397B-A17B & \cmark & \Best{75.94} & \Second{50.62} & \Best{78.24} & \Best{68.27} & \Best{94.54} & \Best{74.49} & \Best{35.77} \\
Qwen3.5-35B-A3B & \cmark & \Second{75.10} & \Best{51.55} & \Second{69.96} & \Second{65.54} & \Second{92.33} & \Second{70.84} & \Second{33.80} \\
KimiK2.5 & \xmark & \Third{71.27} & \Third{41.95} & \Third{65.02} & \Third{59.41} & \Third{89.94} & \Third{62.79} & \Third{25.51} \\
Qwen3VL-32B-I & \xmark & 51.35 & 10.91 & 63.25 & 41.84 & 78.70 & 41.36 & 5.46 \\
InternVL3.5-38B & \xmark & 37.81 & 32.63 & 46.23 & 38.96 & 70.43 & 36.54 & 11.35 \\
Qwen3VL-8B-I & \xmark & 31.14 & 15.06 & 45.46 & 30.55 & 64.49 & 23.71 & 3.49 \\
InternVL3.5-8B & \xmark & 26.30 & 28.34 & 29.52 & 28.05 & 60.15 & 20.18 & 3.81 \\
InternVL3-38B & \xmark & 30.31 & 20.12 & 32.81 & 27.74 & 59.04 & 21.24 & 2.96 \\
KimiVL-A3B-T & \cmark & 30.17 & 25.52 & 24.98 & 26.91 & 62.17 & 16.23 & 2.17 \\
Qwen3VL-32B-T & \cmark & 25.05 & 25.38 & 25.72 & 25.38 & 58.76 & 15.40 & 1.91 \\
LLaVA-CriticR1-7B & \cmark & 23.81 & 26.90 & 25.36 & 25.36 & 58.22 & 16.17 & 1.73 \\
Qwen2.5VL-7B-I & \xmark & 27.22 & 26.22 & 24.17 & 25.87 & 57.93 & 17.70 & 2.11 \\
GLM4.6V-Flash & \xmark & 21.78 & 14.71 & 39.58 & 25.36 & 55.64 & 17.36 & 3.04 \\
Qwen3VL-4B-T & \cmark & 27.81 & 24.85 & 22.95 & 25.20 & 58.38 & 15.58 & 1.64 \\
Qwen3VL-8B-T & \cmark & 24.65 & 25.38 & 24.59 & 24.87 & 57.53 & 15.45 & 1.64 \\
ThinkLiteVL-7B & \cmark & 24.36 & 25.79 & 23.58 & 24.58 & 57.80 & 14.60 & 1.33 \\
MiMoVL-7B-RL & \xmark & 20.84 & 23.73 & 28.80 & 24.46 & 53.91 & 16.70 & 2.76 \\
Qwen3VL-4B-I & \xmark & 20.38 & 19.72 & 32.48 & 24.19 & 54.30 & 15.51 & 2.76 \\
KimiVL-A3B-I & \xmark & 21.09 & 26.78 & 23.13 & 23.67 & 52.23 & 15.60 & 2.75 \\
MiMoVL-7B-SFT & \xmark & 23.41 & 20.51 & 25.77 & 23.23 & 52.86 & 14.60 & 2.24 \\
Molmo2-8B & \xmark & 24.19 & 20.70 & 24.73 & 23.21 & 53.50 & 14.25 & 1.88 \\
\textbf{Avg.} & & 34.00 & 26.54 & 36.97 & 32.51 & 63.95 & 26.49 & 7.15 \\
\specialrule{0.55pt}{0.2em}{0.2em}
\Baseline{Human\textsuperscript{\(\star\)}} & \Baseline{--} & \Baseline{96.00} & \Baseline{98.00} & \Baseline{97.00} & \Baseline{97.00} & \Baseline{100.00} & \Baseline{99.00} & \Baseline{92.00} \\
Random & -- & 25.00 & 25.00 & 25.00 & 25.00 & 57.81 & 15.62 & 1.56 \\
\bottomrule
\end{tabularx}
\vspace{-0.8em}
\end{table*}

\section{\bench}
\label{sec:bench}

\subsection{Overview}

A video \(V\) is defined over the time interval \((0{,}T)\). We use \(C_{a{,}b}\) to denote the clip spanning interval \((a{,}b)\). For a target interval \((s{,}e)\), the video can be decomposed into the beginning \(B=C_{0{,}s}\), the middle \(M=C_{s{,}e}\), and the ending \(E=C_{e{,}T}\):
\[
V=[C_{0{,}s} \mid C_{s{,}e} \mid C_{e{,}T}]=[B \mid M \mid E].
\]

\noindent \bench instantiates the missing-middle video cloze task. Given the visible context \((B{,}E)\) and a candidate set \(\mathcal{Y}_d\), the model must select the clip that correctly fills the missing interval. For each dimension \(d\), the candidate set contains the middle \(M\) and three distractors:
\[
\mathcal{Y}_d=\{M{,}D_d^1{,}D_d^2{,}D_d^3\}{,} \quad d\in\{S{,}A{,}P\}.
\]

\noindent Distractors are designed around three dimensions: Semantic (S), Alignment (A), and Progression (P).

\subsection{Data Sources}
We build \bench from seven public sources: CaReBench~\citep{xu2026carebench}, Daily-Omni~\citep{zhou2025dailyomni}, EgoLife~\citep{yang2025egolife}, FAVOR-Bench~\citep{tu2025favorbench}, LVD-2M~\citep{xiong2024lvd2m}, MiraData~\citep{ju2024miradata}, and Video Thinking Test~\citep{zhang2025videott}. Long-take sources such as MiraData and LVD-2M and egocentric sources including EgoLife and FAVOR-Bench provide temporally continuous videos that reduce scene-change cues, while the remaining sources broaden the activity domains. Detailed dataset descriptions are provided in Appendix~\ref{sec:source_dataset_overview}.

\subsection{Dataset Construction}
We construct \bench with a pipeline that filters videos efficiently before synthesizing distractors.

\subsubsection{Video Filtering}

\noindent\textbf{Duration and LLM Filtering.}
We first rely on source metadata to retain videos between 12 and 90 seconds. We leverage temporally dense captions provided by sources such as LVD-2M and MiraData to screen videos before loading the raw files. We then employ the reasoning model GPT-o3~\citep{openai-o3} to assess whether the events are temporally coherent and whether the surrounding context can constrain a missing middle. Videos that are unsuitable for cloze formatting are excluded.

\vspace{2mm}

\noindent\textbf{Quality and Motion Filtering.}
We remove clips with a bitrate below 200 kbps or sharpness below 30. For videos that pass quality checks, we sample a gap from the central 50\% of the timeline, with length set to 20\%--40\% of the full duration. This leaves context on both sides and avoids trivial and underconstrained boundary cases. We validate each gap with Farneb\"ack optical flow~\citep{farneback2003two}, requiring average flow magnitude above 1.0 and resampling up to three times.

These stages preserve video quality while reducing unnecessary loading. After this pipeline, \bench retains 1,521 videos. Table~\ref{tab:sources} summarizes the source and duration statistics. Appendix~\ref{sec:filtering_statistics} provides detailed filtering statistics, and Appendix~\ref{sec:filtered_video_human_validation} reports human validation of filtered-video quality.

\subsubsection{Distractor Generation}
We construct three distractors per dimension. Formal mathematical definitions are in Appendix~\ref{sec:distractor_formalization}.

\noindent\textbf{Semantic} focuses on the specific content that the missing clip needed to connect. Each distractor is a same-duration clip, \(D_S=C_{u{,}u+\ell}\), where \((u{,}u+\ell)\) does not overlap \((s{,}e)\). Distinguishing them requires identifying the event implied by the context and the core activity of the video.

\vspace{2mm}

\noindent\textbf{Alignment} focuses on the temporal compatibility of the missing clip when the content remains plausible. Each distractor is a nearby clip, \(D_A=C_{s'{,}e'}\), where \((s'{,}e')\) shifts or expands \((s{,}e)\). Advanced and Deferred place the candidate before or after the target, while Expanded widens it to include adjacent context from both. Distinguishing them requires sensitivity to temporal boundaries.

\vspace{2mm}

\noindent\textbf{Progression} focuses on event development of the missing clip when both content and coarse temporal compatibility remain plausible. We use Reversed, Reordered, and Repeated variants: Reversed plays the answer segment backward, Reordered disrupts the local order of subevents, and Repeated duplicates a short subsegment. Distinguishing them requires fine-grained perception of event direction, subevent order, and action continuity.

\begin{figure*}[t]

  \centering
  \captionsetup[subfigure]{skip=2pt} 
  \begin{subfigure}[b]{0.35\textwidth}
    \centering
    \includegraphics[width=\linewidth]{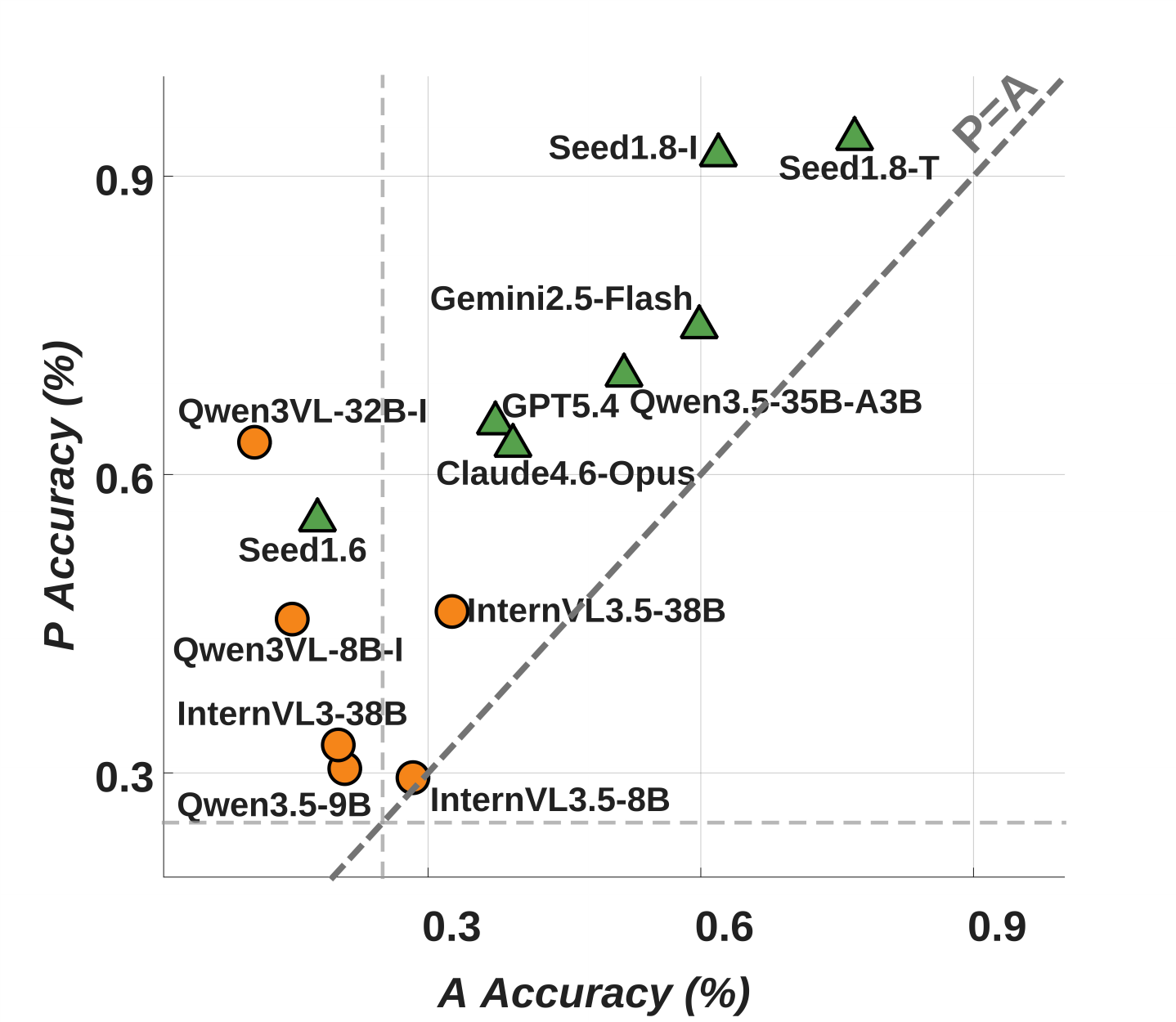}
    \caption{Progression vs. Alignment}
  \end{subfigure}\hspace{-0.03\textwidth}
  \begin{subfigure}[b]{0.35\textwidth}
    \centering
    \includegraphics[width=\linewidth]{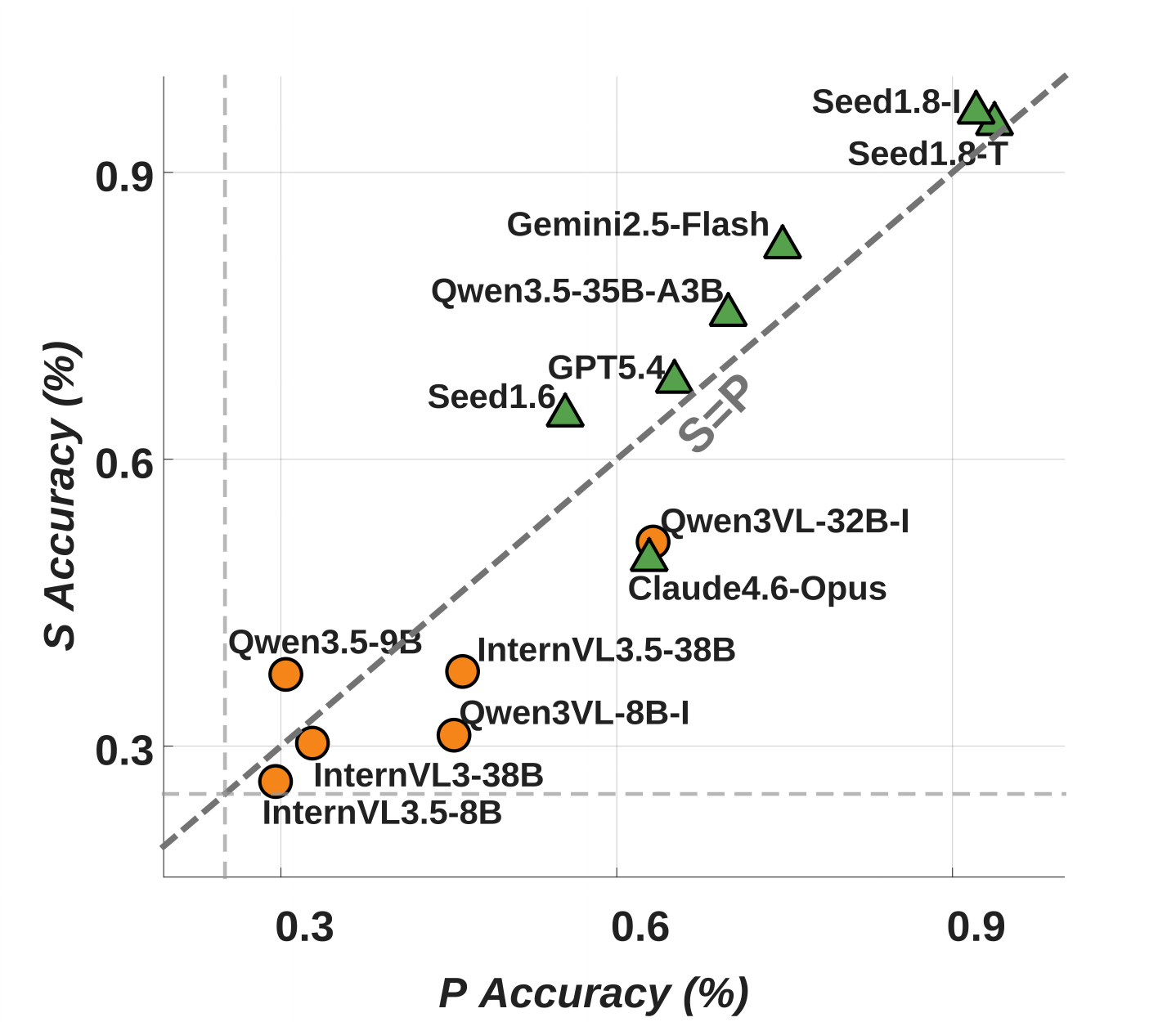}
    \caption{Semantic vs. Progression}
  \end{subfigure}\hspace{-0.03\textwidth}
  \begin{subfigure}[b]{0.35\textwidth}
    \centering
    \includegraphics[width=\linewidth]{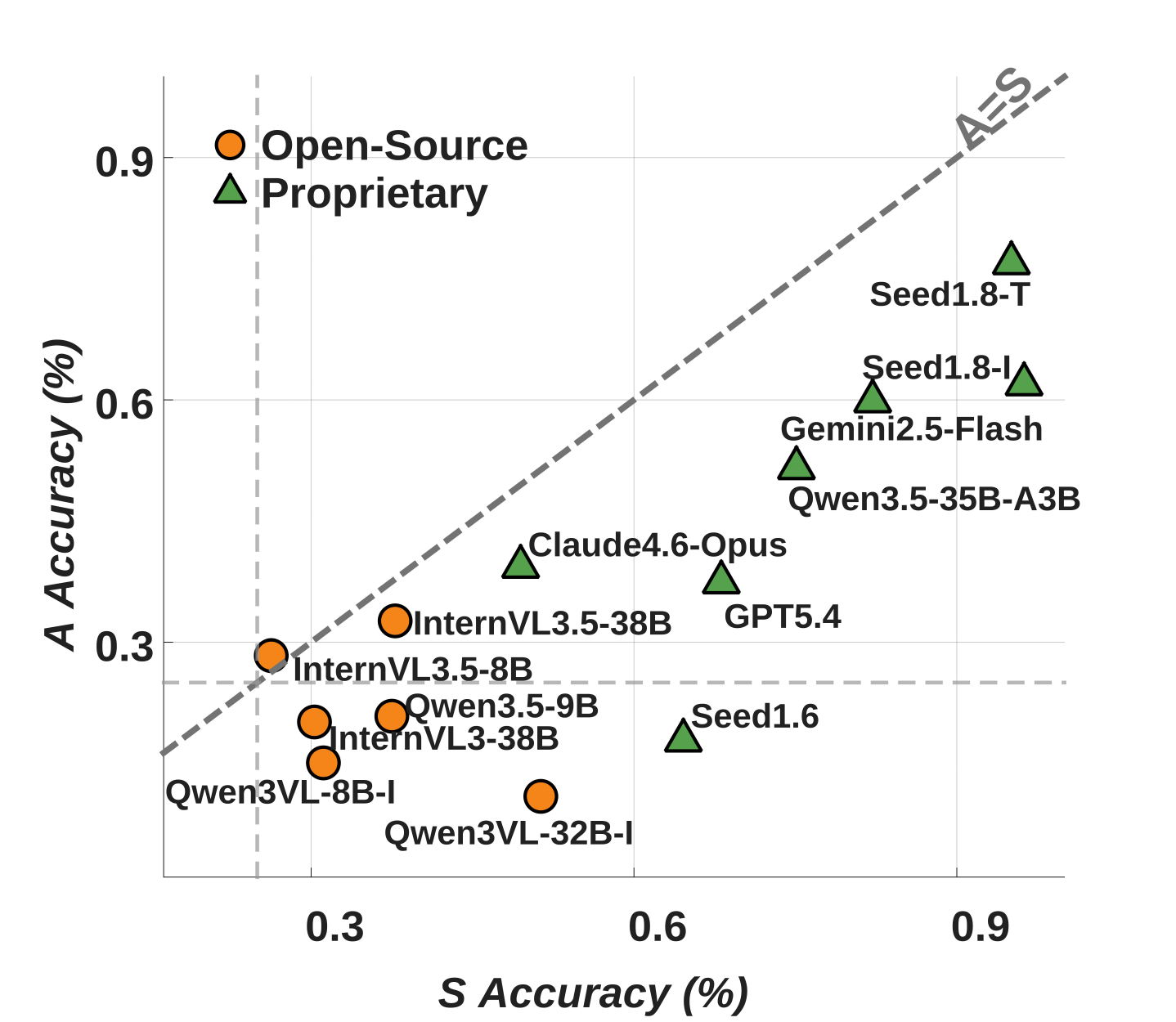}
    \caption{Alignment vs. Semantic}
  \end{subfigure}
  \vspace{-0.5em}
  \caption{Pairwise comparison of dimension-level accuracy on \bench across leading models.}
  \label{fig:sap}
  \vspace{-1.3em}
\end{figure*}

\section{Experiments}

\subsection{Setup}
\label{sec:exp_setup}

\subsubsection{Models}
We evaluate 10 proprietary and 21 open-source models. For models with different versions, we use the suffix T for thinking and I for instruct. Proprietary models are Qwen3.5-Plus~\citep{qwen2026qwen35}, Seed1.8~\citep{bytedance2025seed18}, Gemini2.5-Pro, Gemini2.5-Flash~\citep{gemini2025gemini25}, GPT5.4~\citep{openai2026gpt54}, Claude4.6-Sonnet, Claude4.6-Opus~\citep{anthropic2026opus46}, Gemini3-Flash~\citep{google2025gemini3flashpreview}, Seed1.6~\citep{bytedance2025seed16}, and Grok4.1~\citep{xai2025grok41}. Open-source models are Qwen3.5-397B-A17B, Qwen3.5-35B-A3B~\citep{qwen2026qwen35}, KimiK2.5~\citep{kimiteam2026k25}, Qwen3VL-32B-I, Qwen3VL-32B-T, Qwen3VL-8B-T, Qwen3VL-8B-I, Qwen3VL-4B-T, Qwen3VL-4B-I~\citep{bai2025qwen3vl}, InternVL3.5-38B, InternVL3.5-8B~\citep{wang2025internvl35}, InternVL3-38B~\citep{zhu2025internvl3}, KimiVL-A3B-T, KimiVL-A3B-I~\citep{kimiteam2025kimivl}, LLaVA-CriticR1-7B~\citep{wang2025llavacriticr1}, Qwen2.5VL-7B-I~\citep{bai2025qwen25vl}, GLM4.6V-Flash~\citep{zai2025glm46v}, ThinkLiteVL-7B~\citep{wang2025thinklitevl}, MiMoVL-7B-RL, MiMoVL-7B-SFT~\citep{yue2025mimovl}, and Molmo2-8B~\citep{clark2026molmo2}.

\subsubsection{Configurations}
By default, we uniformly sample 16 frames from each clip, yielding 96 frames per dimension across six clips. Appendix~\ref{sec:bin_centered_sampling} details the sampling rule. To avoid exact boundary-frame shortcuts, we also conduct a boundary sanity check in Appendix~\ref{sec:boundary_sanity_check}. Open-source models are deployed on A6000 GPUs using vLLM~\citep{kwon2023vllm}.

\subsubsection{Representative Models and Subsets}
For in-depth analysis, we use four representative models: Gemini2.5-Pro, Seed1.6, Qwen3.5-397B-A17B, and Qwen3VL-8B-I. We also construct two auxiliary subsets. TempCloze-Mixed contains a randomly shuffled subset of 300 videos, each with candidates from the ground-truth and S, A, and P. TempCloze-Hard contains the 150 instances with the largest number of errors across models.

\begin{figure*}[t]
  \centering
  \includegraphics[width=0.98\textwidth]{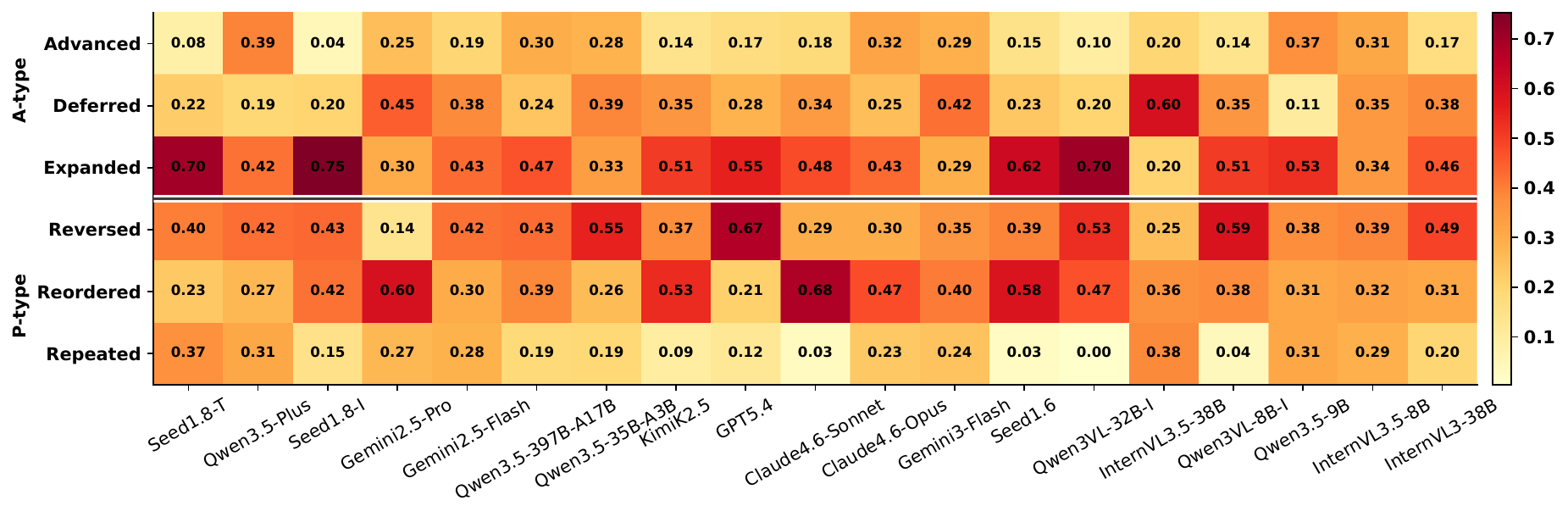}
  \vspace{-1.1em}
  \caption{Error proportions for alignment and progression distractors. Darker cells indicate higher error rates.}
  \label{fig:failure_heatmap}
  \vspace{-1.1em}
\end{figure*}

\subsection{Main Results}
\label{sec:main_results}

Table~\ref{tab:main_results} presents the main results for both proprietary and open-source models.
We provide the full leaderboard in Figure~\ref{fig:top_metrics} and report source-level statistics for Gemini2.5-Pro and Qwen3.5-35B-A3B in Appendix~\ref{sec:source_statistics}.

\vspace{0.35em}
\noindent\textbf{Dimension accuracy.}
On Semantic, proprietary models average 70.73\%, led by Seed1.8 at 96.25\%.
Open-source models average 34.00\%.
Nevertheless, Qwen3.5-397B reaches 75.94\%, showing that the strongest open-source systems can approach proprietary-level Semantic performance.
A similar group-level gap appears on Progression, where proprietary models average 67.72\% and open-source models average 36.97\%. Qwen3.5-397B reaches 78.24\%, indicating that stronger open-source models can identify local event order.
Alignment is substantially harder.
Proprietary models drop to 48.13\% on average, while open-source models drop to 26.54\%.
Even the strongest scores, 76.92\% for Seed1.8 and 51.55\% for Qwen3.5-35B, are well below the 98.00\% Human Baseline.
\textbf{\textit{Alignment is the primary bottleneck for visual temporal reasoning in current Video-LLMs.}}
Models are better at recognizing plausible event content and local progression than at judging exactly when an event should occur relative to the observed endpoints.

To further examine dimension-level behavior, Figure~\ref{fig:sap} compares leading models across pairs.
Two panels involving Alignment reinforce the pattern raised, as most models score lower on Alignment than the paired dimension.
A notable observation emerges from the S--P panel.
Open-source models have low absolute scores but perform better on Progression than Semantic, suggesting that local motion continuity can support a basic Progression score.
By contrast, proprietary models sit at higher scores and favor Semantic over Progression, indicating that high Progression accuracy requires stronger temporal understanding.
We report joint accuracy for dimension pairs in Appendix~\ref{sec:joint_accuracy}.

\begin{table}[t]
\centering
\caption{TempCloze-Mixed error attribution (\%). The Alignment and Progression options are each randomly sampled from one of three designed variants.}
\vspace{-0.5em}
\label{tab:mixed_error}
\footnotesize
\setlength{\tabcolsep}{4pt}
\renewcommand{\arraystretch}{1.16}
\begin{tabular}{@{}l@{\hspace{0.45em}}cccc@{}}
\toprule
\textbf{Model} & \textbf{Acc.} & \textbf{S} & \textbf{A} & \textbf{P} \\
\midrule
Gemini2.5-Pro & 76.3 & 2.7 & \textbf{19.7} & 1.3 \\
Seed1.6 & 68.9 & 7.0 & \textbf{14.7} & 9.4 \\
Qwen3.5-397B-A17B & 62.0 & 9.0 & \textbf{20.3} & 8.7 \\
Qwen3VL-8B-I & 24.0 & 29.3 & \textbf{37.3} & 9.3 \\
\bottomrule
\end{tabular}
\vspace{-1.5em}
\end{table}

\noindent\textbf{Cumulative accuracy.}
Cumulative results shift the view from dimensions to full-video understanding.
We interpret $\geq\!1$ as limited understanding, $\geq\!2$ as sufficient understanding, and 3/3 as complete understanding.
Proprietary models show a meaningful level of sufficient understanding, averaging 65.12\% on $\geq\!2$ and reaching 95.40\% for the best model.
However, complete understanding is harder.
The proprietary average drops to 33.24\% on 3/3, and even the best model reaches only 70.81\%, far below the 92.00\% Human Baseline.
Open-source models show a more limited level of sufficient understanding. Even the strongest one reaches only 74.49\% on $\geq\!2$ and 35.77\% on 3/3, indicating that open-source models have not yet reached sufficient understanding.
Overall, current Video-LLMs remain short of complete video understanding that requires robust visual temporal reasoning.
We provide a video-level breakdown in Appendix~\ref{sec:video_level_breakdown}.

\begin{table}[bth]
\vspace{-0.5em}
\centering
\caption{Effect of candidate order permutation on TempCloze-Hard across representative models. Disp. denotes accuracy dispersion. FR denotes Flip Rate, and CFR denotes Clip Flip Rate. All values are percentages.}
\vspace{-0.5em}
\label{tab:perm_consistency}
\footnotesize
\setlength{\tabcolsep}{1.2pt}
\renewcommand{\arraystretch}{0.93}
\resizebox{\columnwidth}{!}{%
\begin{tabular}{@{}>{\raggedright\arraybackslash}p{0.190\columnwidth}*{6}{>{\centering\arraybackslash}p{0.135\columnwidth}}@{}}
\toprule
\textbf{Model} & \textbf{Dim.} & \textbf{Mean} & \textbf{Range} & \textbf{Disp.} & \textbf{FR} & \textbf{CFR} \\
\midrule
\multirow{4}{*}{\makecell[l]{Gemini2.5\\-Pro}}
& S & 70.0 & 9.3 & 3.9 & 30.2 & 36.9 \\
& A & 41.2 & 6.0 & 2.6 & 41.9 & 64.6 \\
& P & 49.7 & 10.0 & 3.7 & 48.4 & 64.8 \\
& \textbf{All} & \textbf{53.6} & \textbf{3.8} & \textbf{1.5} & \textbf{40.2} & \textbf{55.4} \\
\midrule
\multirow{4}{*}{Seed1.6}
& S & 76.8 & 6.0 & 2.4 & 18.8 & 22.0 \\
& A & 33.7 & 1.3 & 0.6 & 34.9 & 48.4 \\
& P & 81.8 & 6.7 & 2.4 & 24.3 & 26.8 \\
& \textbf{All} & \textbf{64.1} & \textbf{2.7} & \textbf{0.9} & \textbf{26.0} & \textbf{32.4} \\
\midrule
\multirow{4}{*}{\makecell[l]{Qwen3.5\\-397B\\-A17B}}
& S & 61.1 & 8.8 & 3.2 & 31.2 & 40.2 \\
& A & 39.1 & 10.1 & 3.9 & 36.6 & 56.9 \\
& P & 64.3 & 10.1 & 4.1 & 36.4 & 44.2 \\
& \textbf{All} & \textbf{54.8} & \textbf{3.1} & \textbf{1.2} & \textbf{34.7} & \textbf{47.2} \\
\midrule
\multirow{4}{*}{\makecell[l]{Qwen3VL\\-8B-I}}
& S & 7.4 & 4.7 & 1.7 & 11.5 & 51.2 \\
& A & 11.7 & 6.0 & 2.3 & 20.6 & 66.1 \\
& P & 37.4 & 4.7 & 1.9 & 45.5 & 64.8 \\
& \textbf{All} & \textbf{18.9} & \textbf{1.8} & \textbf{0.8} & \textbf{25.9} & \textbf{60.7} \\
\bottomrule
\end{tabular}
}
\vspace{-1.5em}
\end{table}

\subsection{Error Pattern Analysis}

\subsubsection{Dimension-Specific Errors}
Because Alignment and Progression use controlled temporal edits rather than random clips, Figure~\ref{fig:failure_heatmap} analyzes their error patterns.
The resulting errors are structured rather than random.
For Alignment, Expanded is the most prominent signature. The selected clip contains the correct event content but extends both before and after the target interval. For example, Seed1.8-I selects Expanded in 75\% of its Alignment errors. For Progression, Reversed is the strongest failure mode. GPT-5.4 selects Reversed in 67\% of its Progression errors. These two errors share visual plausibility, as models select clips whose content still appears relevant in the surrounding video while failing at temporal boundaries or at the directional order of event development.

\begin{figure}[t]
  \centering
  \includegraphics[width=\columnwidth]{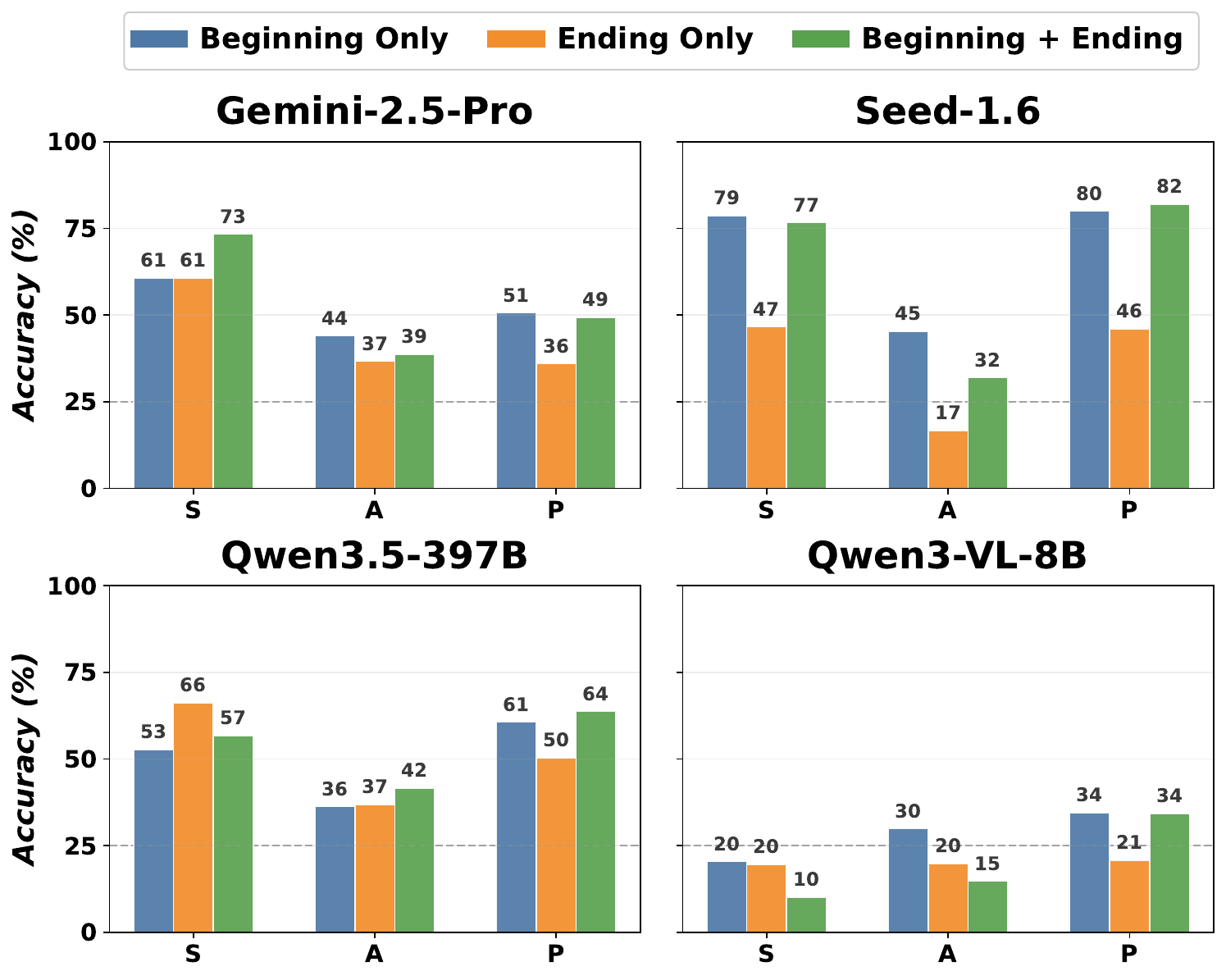}
  \vspace{-2em}
  \caption{Effect of context direction ablation on TempCloze-Hard across representative models.}
  \label{fig:context_ablation}
  \vspace{-1.5em}
\end{figure}

\vspace{-2mm}
\subsubsection{Mixed-Dimension Errors}
Table~\ref{tab:mixed_error} reports accuracy together with the share of each distractor dimension on TempCloze-Mixed. Across models, Alignment receives the highest error share.
This makes Alignment the most distracting dimension, echoing the finding raised in Section~\ref{sec:main_results} that Alignment is the primary bottleneck. A counterintuitive pattern is that Semantic can be more distracting than Progression, even though models generally achieve slightly higher Semantic accuracy than Progression accuracy in Table~\ref{tab:main_results}. This suggests that incorrect progression is easier to reject than an incorrect event. Beyond dimension shares, the accuracy drop indicates that competition across dimensions is harder than evaluating a single dimension. Although this setting reduces the diagnostic separation within individual dimensions, it shows that models have substantial room to improve in visual temporal reasoning, and reveals failures hidden by marginal accuracy alone.

\begin{figure}[t]
  \centering
  \includegraphics[width=\columnwidth]{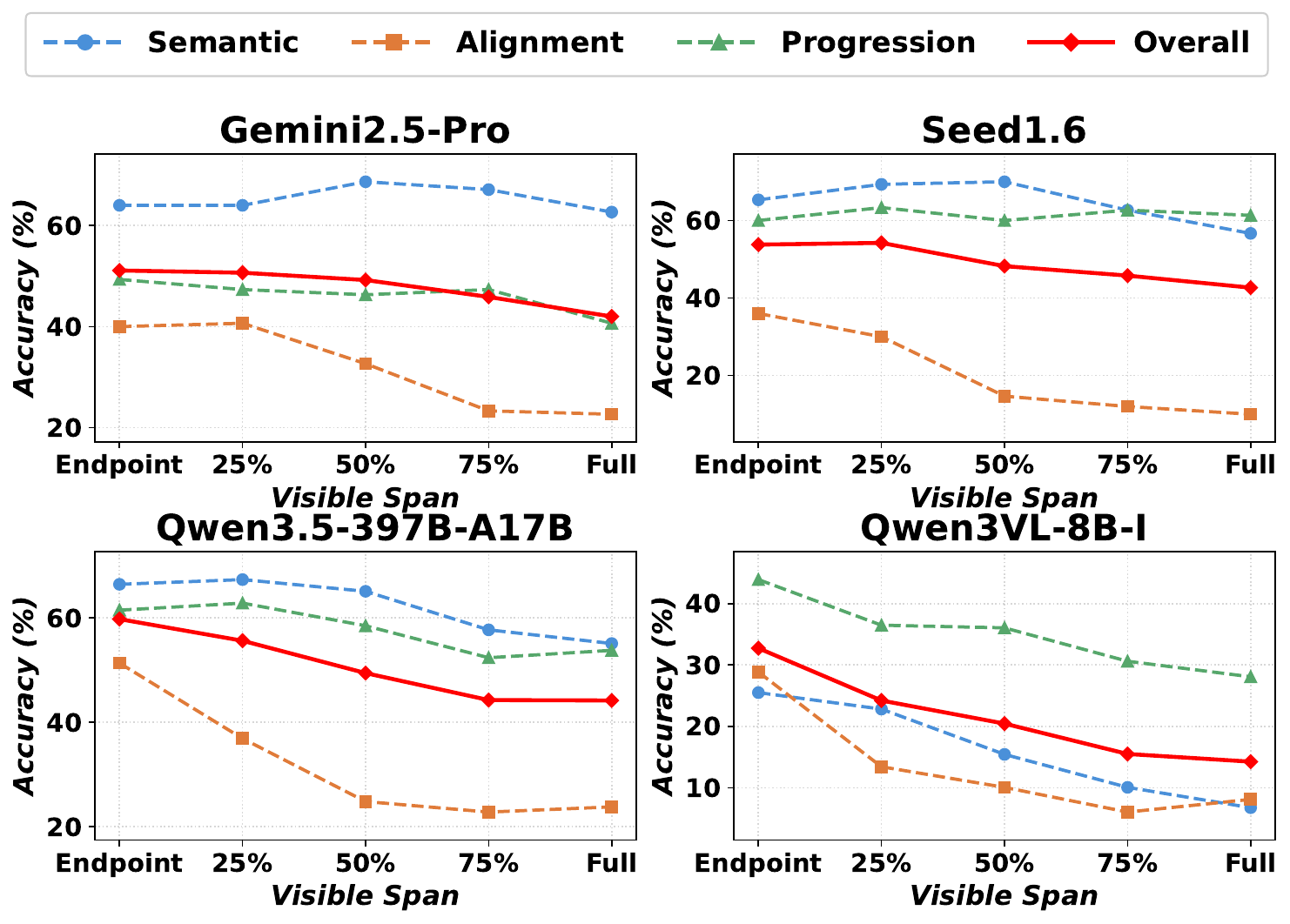}
  \vspace{-2em}
  \caption{Effect of visible span scaling on TempCloze-Hard across representative models.}
  \label{fig:visible_span}
  \vspace{-1.5em}
\end{figure}

\vspace{-2mm}
\subsection{Behavioral Sensitivity Analysis}
On TempCloze-Hard, we examine how multiple factors affect representative models.

\subsubsection{Candidate Order Permutation}

As shown in Table~\ref{tab:perm_consistency}, we evaluate the effect of candidate order by testing four candidate permutations per instance at temperature 0. FR is the share of instances for which the correctness changes at least once and CFR is the share for which the selected clip changes at least once.
Mean accuracy changes little, with variance below 4 points for every model, but both instability measures remain substantial: CFR ranges from 32.4\% to 60.7\%, and FR ranges from 25.9\% to 40.2\%.
This indicates that stable accuracy can hide unstable choices.
Another consistent pattern is that Alignment and Progression are generally less stable than Semantic, suggesting that visual temporal reasoning is sensitive to candidate order when candidates are visually similar but temporally distinct.
This is expected because content differences are salient, whereas timing and progression differences are subtler.

\vspace{-1mm}

\subsubsection{Context Direction Ablation}

Figure~\ref{fig:context_ablation} examines which part of the visible context plays a stronger role in model decisions.
We observe that ending-only is often the weakest, while beginning-only frequently approaches or exceeds the performance of both contexts. For example, Seed-1.6 reaches 80\% on Progression with beginning-only, close to 82\% with both, but drops to 46\% with ending-only.
The asymmetry suggests that models rely more on forward continuation cues than backward inference. This may reflect forward ordered pretraining traces, which bias inference in the same direction. Consequently, jointly considering information from both directions remains a difficulty for visual temporal reasoning.

\vspace{-1mm}

\subsubsection{Visible Span Scaling}

Figure~\ref{fig:visible_span} studies how performance changes as the visible span expands from endpoint frames to full context.
Alignment drops sharply for all models, since its evidence lies near the missing gap boundaries and longer spans dilute these cues.
For Semantic and Progression, stronger models stay stable or improve slightly at short spans before declining mildly, while weaker models degrade much more clearly.
This pattern suggests that longer context is double-edged: it adds temporal evidence for event-level reasoning but can distract models from the most diagnostic frames.
Current Video-LLMs still need stronger mechanisms for integrating long-range context and selecting critical frames.

\vspace{-1mm}

\subsubsection{Sampling Density Scaling}

\begin{figure}[t]
    \centering
    \includegraphics[width=\columnwidth]{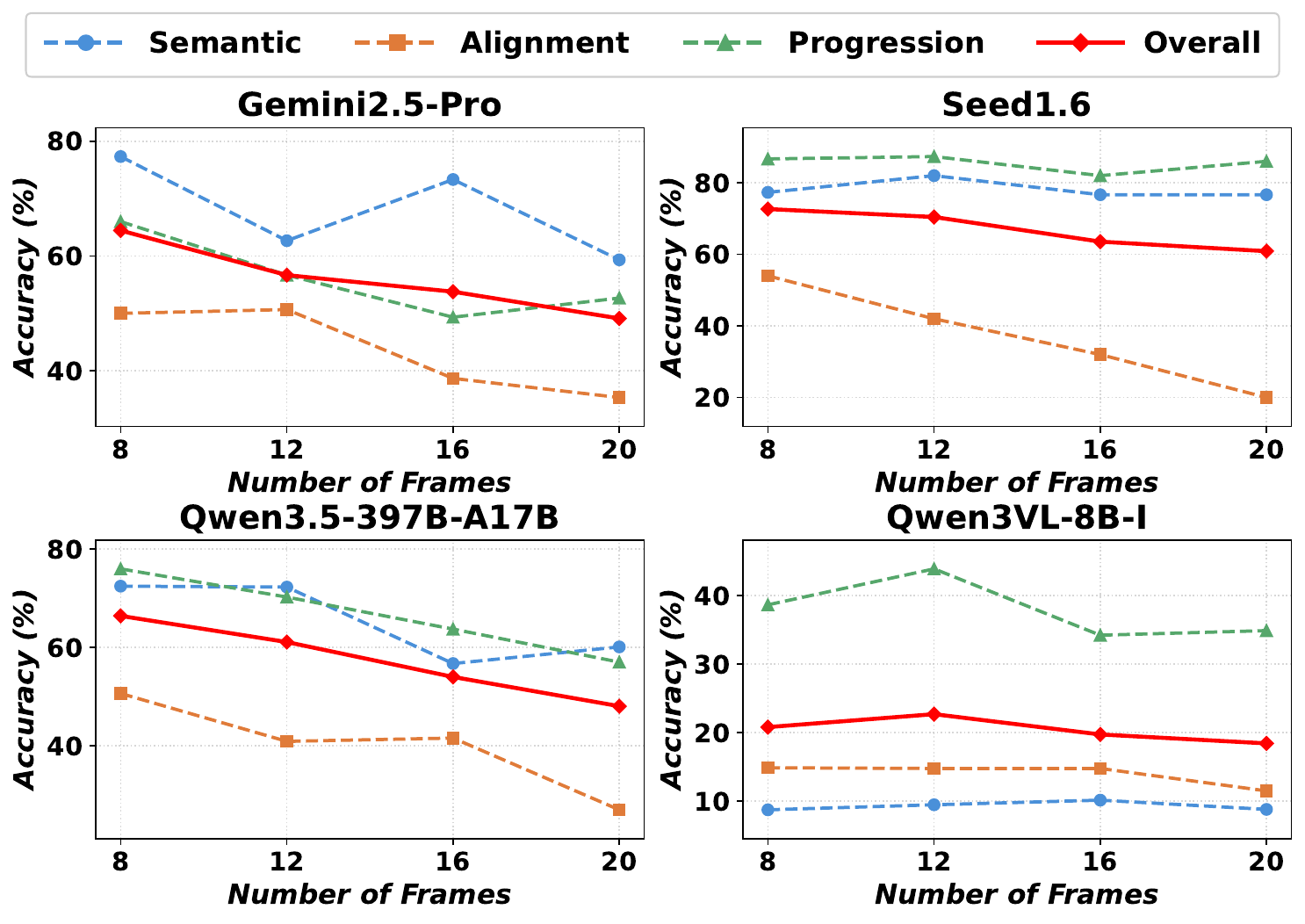}
    \vspace{-2em}
    \caption{Effect of sampling density scaling on TempCloze-Hard across representative models.}
    \label{fig:frames_dims}
    \vspace{-1.5em}
\end{figure}

Figure~\ref{fig:frames_dims} studies how performance changes as sampling density increases from 8 to 20 frames per clip.
The pattern is similar to visible span scaling: stronger models are more likely to preserve or slightly improve Semantic and Progression, while Alignment declines across models.
This suggests that higher sampling density does not automatically provide richer evidence for Video-LLMs.
Instead, dense similar frames can add confusion and reduce attention to temporally decisive cues.
Exploring video native architectures that do not treat videos as sequences of images may be a promising direction for exploiting dense temporal evidence.

\vspace{-1mm}

\subsubsection{Test-Time Scaling}
\label{sec:test_time_scaling}

\begin{figure}[t]
    \centering
    \includegraphics[width=\columnwidth]{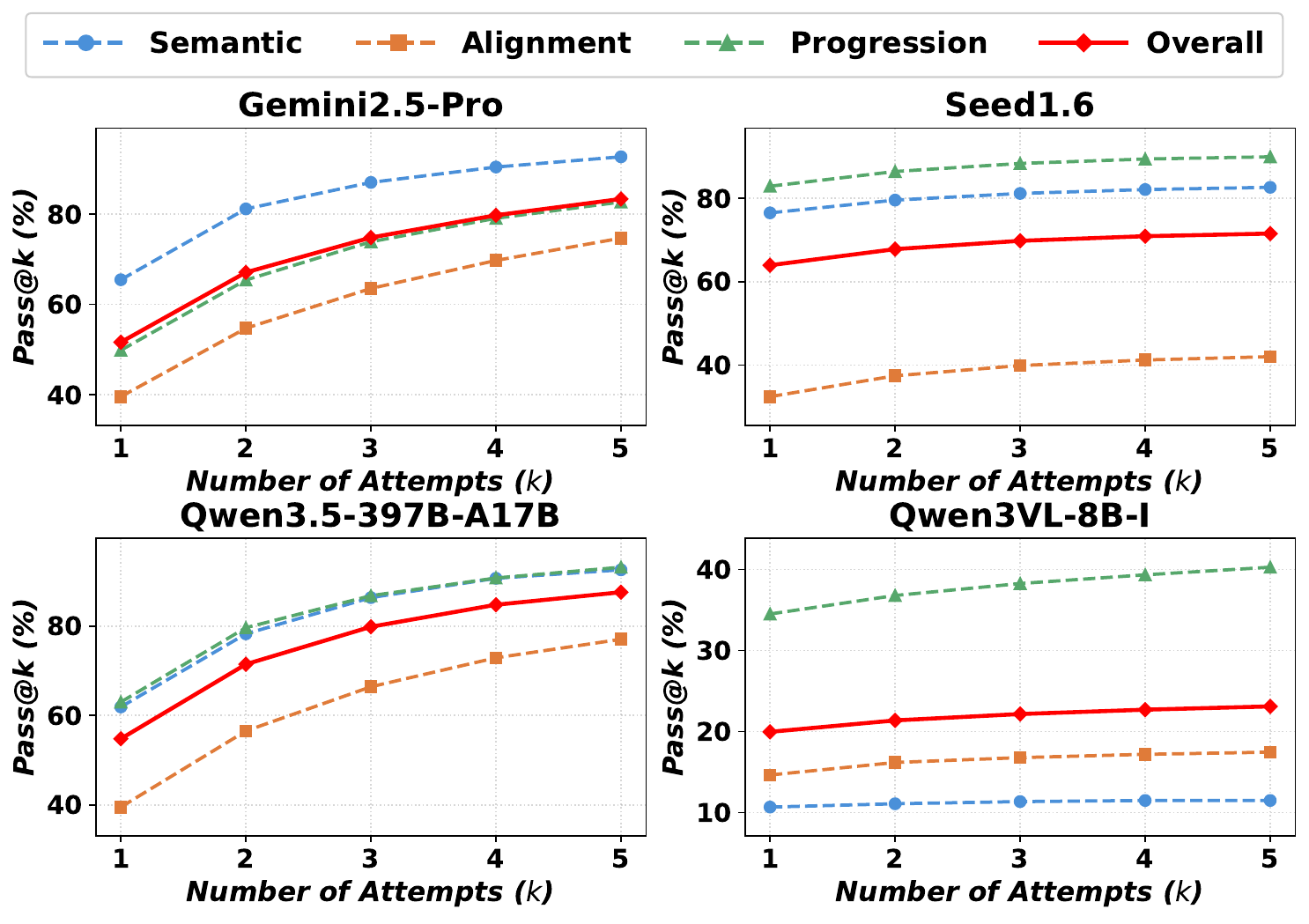}
    \vspace{-2em}
    \caption{Effect of test-time scaling on TempCloze-Hard across representative models.}
    \label{fig:passk_dims}
    \vspace{-1.5em}
\end{figure}

Figure~\ref{fig:passk_dims} explores the test-time performance upper bound at temperature 0.7.
Pass@\(k\) counts an instance as solved if any of \(k\) sampled attempts is correct.
All models improve with more attempts, but the gain varies substantially.
Gemini2.5-Pro and Qwen3.5-397B-A17B show the largest gains, with Overall pass@\(k\) rising by roughly 30 points from \(k=1\) to \(k=5\).
This suggests that their predictions are more unstable, consistent with Table~\ref{tab:perm_consistency}, where their Disp. values are higher (1.5 and 1.2) than those of Seed1.6 and Qwen3VL-8B-I (0.9 and 0.8).
However, extra attempts do not change the relative ordering among dimensions: Alignment remains the lowest as the bottleneck while Semantic and Progression vary by model.

\vspace{-2mm}
\section{Conclusion}
\vspace{-2mm}
\bench evaluates visual temporal reasoning by requiring models to identify the missing middle from surrounding clips, instead of reasoning over natural-language descriptions.
Across 1,521 videos, it builds same-source candidates along three dimensions: \textit{Semantic} targets what event should happen, \textit{Alignment} probes when it should occur, and \textit{Progression} tests how it should unfold.
Because candidates share scenes and objects, \bench limits appearance shortcuts and tests temporal evidence among visually similar options.

Results expose an uneven boundary in Video-LLMs.
Many models identify Semantic content and event Progression, but temporal Alignment remains the bottleneck: models struggle to place the missing middle precisely between endpoints, while humans solve the task reliably.
Error and sensitivity analyses reveal a finer-grained picture.
Within dimensions, failures often choose clips with plausible content but incompatible temporal structure. Across dimensions, Alignment is the most confusing. Models remain sensitive to presentation order and context direction, showing that current gains do not reflect stable reasoning. Even with test-time scaling,the relative ordering is largely unchanged and keeps Alignment as the hardest.

Looking forward, Video-LLMs need video native mechanisms that integrate both temporal directions, select key evidence, and represent temporal structure explicitly.
Future benchmarks should focus on visual temporal reasoning in longer videos while controlling linguistic shortcuts.


\section*{Limitations}

TempCloze is a controlled diagnostic benchmark for missing-middle identification, rather than a complete measure of all video understanding abilities. Its design focuses on visually grounded temporal fit between the observed beginning and ending clips. It therefore does not directly cover open-ended generation, dialogue-based narrative understanding, audio-grounded inference, or unconstrained future prediction. The fixed candidate format also means that performance should be interpreted with respect to the designed alternatives, rather than as a general measure of free-form video reasoning.

All candidates are drawn from the same source video to reduce appearance shortcuts and keep alternatives visually plausible. This choice makes the task stricter: TempCloze measures exact temporal compatibility, not only broad event plausibility. In particular, Alignment distractors may contain relevant content while still failing to match the removed interval. Such cases are intentional, as they test whether models can identify the correct temporal span instead of selecting a generally plausible continuation.

Our model results should be viewed as a snapshot of current Video-LLMs. APIs, decoding behavior, and video input limits can change over time. TempCloze also draws from public sources that emphasize long-take, egocentric, and fine-grained motion videos, which supports temporal comparison but cannot cover every domain or camera style. Future extensions can broaden video domains and modalities while preserving the same visual candidate-comparison principle.

\section*{LLM Usage Statement}
\label{app:llmuse}
Central to our methodology, multiple Large Language Models (LLMs) served as the subjects of our experiments to test our proposed benchmark. We explicitly state that we have never relied on LLMs to generate core research ideas, methodologies, experimental designs, or conclusions. All technical contributions and analyses presented are the original work of the authors.
\bibliography{custom}
\clearpage
\appendix
\newpage
\begin{figure*}[tbh]
\vspace*{-0.8em}
\centering
\includegraphics[width=0.985\textwidth]{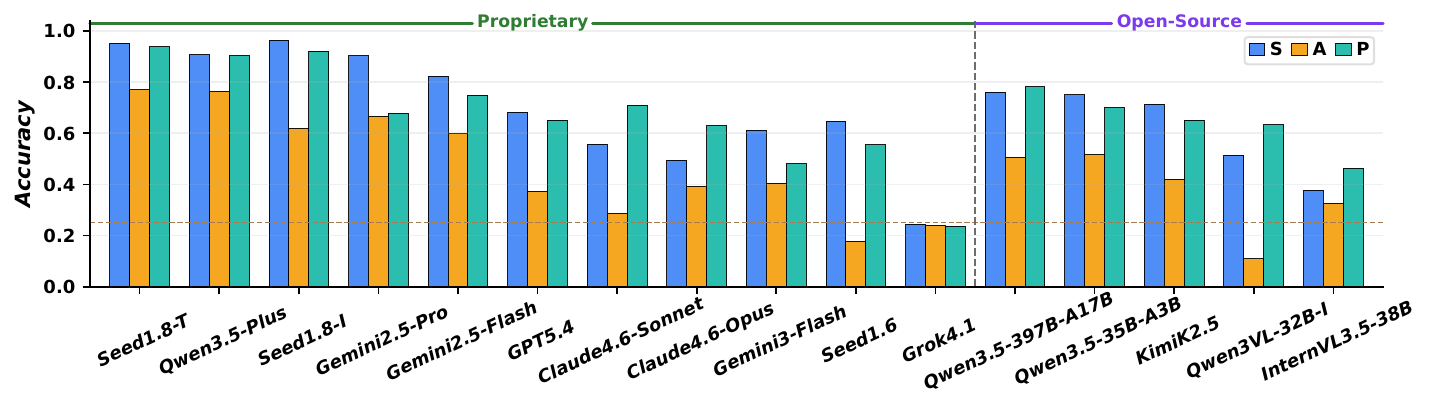}
\vspace{-0.5em}
\caption{Top metrics of proprietary and open-source models on \bench.}
\label{fig:top_metrics}
\end{figure*}

\section{Distractor Formalization}
\label{sec:distractor_formalization}

Using the notation in Section~\ref{sec:bench}, we define the distractors below. All intervals are valid source-video subclips. The ground-truth interval \((s,e)\) lies in the central 50\% of the video and has duration \(\ell=e-s\in[0.2T,0.4T]\). Advanced and Deferred shift it by \(\ell/2\), while Expanded extends it by \(\ell/2\) on each side; hence \(\delta=\delta_1=\delta_2=\ell/2\in[0.1T,0.2T]\), preserving context on both sides.

\paragraph{Semantic}
distractors are same-duration clips outside the missing interval:
\[
\begin{array}{rcl}
D_S^i &=& C_{u_i{,}u_i+\ell}{,}\\[0.45em]
(u_i{,}u_i+\ell)\cap(s{,}e) &=& \emptyset.
\end{array}
\]

\paragraph{Alignment}
distractors stay near the target but perturb its boundaries:
\[
\begin{array}{rcl}
D_A^{\mathrm{advanced}} &=& C_{s-\delta{,}e-\delta}{,}\\[0.45em]
D_A^{\mathrm{deferred}} &=& C_{s+\delta{,}e+\delta}{,}\\[0.45em]
D_A^{\mathrm{expanded}} &=& C_{s-\delta_1{,}e+\delta_2}.
\end{array}
\]
The sampling procedure retains only perturbations whose intervals remain inside the source video.

\paragraph{Progression}
distractors alter the target span's internal order. For reorder, \(J_1{,}\ldots{,}J_m\) partitions \((s{,}e)\) and \(\pi\) permutes \(\{1{,}\ldots{,}m\}\); for repeat, \(J\subset(s{,}e)\) is a short subinterval:
\[
\begin{array}{rcl}
D_P^{\mathrm{rev}} &=& \mathrm{reverse}\!\left(C_{s{,}e}\right){,}\\[0.45em]
D_P^{\mathrm{reorder}} &=& \mathrm{concat}_{k=1}^{m} C_{J_{\pi(k)}}{,}\\[0.45em]
D_P^{\mathrm{repeat}}  &=& \mathrm{trim}_{\ell}(C_J \mid C_J \mid \cdots).
\end{array}
\]

\section{Frame Sampling}
\label{sec:frame_sampling}

\subsection{Bin-Centered Sampling}
\label{sec:bin_centered_sampling}

For each input clip, we divide the interval \((a{,}b)\) into \(n\) equal bins and sample each bin center:
\[
t_i = a + \left(i+\frac{1}{2}\right)\frac{b-a}{n}{,} \quad i=0{,}\ldots{,}n-1.
\]
Decoded frames are selected from these timestamps. Unlike endpoint-inclusive sampling, this keeps the first and last samples inside the clip rather than exactly at \(a\) or \(b\). This reduces boundary-frame leakage, since visible context clips meet the missing middle at exact temporal boundaries and some candidates touch or overlap those boundaries.

\subsection{Boundary Sanity Checks}
\label{sec:boundary_sanity_check}

\subsubsection{Exact Boundary-Frame Matches}
\label{sec:exact_boundary_frame_matches}

We test whether each candidate's first or last 1--3 sampled frames exactly match the adjacent context boundary. Across 1,521 videos per dimension and 18,252 candidate clips, Table~\ref{tab:boundary_sanity} shows zero matches.

\begin{table}[ht]
\vspace{-0.5em}
\centering
\caption{Exact boundary-frame matches in sampled candidates.}
\vspace{-0.5em}
\label{tab:boundary_sanity}
\footnotesize
\renewcommand{\arraystretch}{1.08}
\setlength{\tabcolsep}{3.6pt}
\begin{tabular}{@{}l r ccc ccc@{}}
\toprule
\multirow{2}{*}{\textbf{Dimension}} & \multirow{2}{*}{\textbf{Clips}} & \multicolumn{3}{c}{\textbf{Begin-side}} & \multicolumn{3}{c}{\textbf{End-side}} \\
\cmidrule(lr){3-5} \cmidrule(l){6-8}
& & \textbf{1f} & \textbf{2f} & \textbf{3f} & \textbf{1f} & \textbf{2f} & \textbf{3f} \\
\midrule
S & 6084  & 0 & 0 & 0 & 0 & 0 & 0 \\
A & 6084  & 0 & 0 & 0 & 0 & 0 & 0 \\
P & 6084  & 0 & 0 & 0 & 0 & 0 & 0 \\
Overall & 18252 & 0 & 0 & 0 & 0 & 0 & 0 \\
\bottomrule
\end{tabular}
\vspace{-1.5em}
\end{table}

\subsubsection{Edge Baseline}
\label{sec:edge_baseline}

Edge-1 uses each candidate's first and last frame; Edge-4 uses its first and last four. We average DINOv2 similarities between candidate edges and adjacent context frames, then normalize the four scores into probabilities.

Table~\ref{tab:edge_baseline} reports mean ground-truth probabilities for Semantic, Alignment, and Progression; Mean averages them and is not top-1 accuracy. Values only slightly above the uniform 25\% show that edges can occasionally be decisive but are not substantial overall.

\newcommand{\BreakdownBlockRow}[1]{
    \specialrule{0.55pt}{0.2em}{0pt}
    \rowcolor{gray!16}[0pt][0pt]
    \multicolumn{10}{@{}c@{}}{\rule{0pt}{2.45ex}\normalsize\itshape #1}\\
    \specialrule{0.55pt}{0pt}{0.2em}
}
\begin{table*}[tbh]
\centering
\caption{
Joint accuracy and video-level breakdown on \bench. Joint columns count videos where the listed dimensions are all correct; 0/3--3/3 count exact numbers of correct dimensions.
The $S\!+\!A\!+\!P$ column equals 3/3.
}
\label{tab:video_level_breakdown}
\footnotesize
\renewcommand{\arraystretch}{0.95}
\setlength{\tabcolsep}{2.8pt}

\begin{tabularx}{\textwidth}{@{} l @{\hspace{3pt}} c @{\hspace{4pt}} *{8}{Y} @{}}
\toprule
\multicolumn{2}{c}{\textbf{Model}} &
\multicolumn{4}{c}{\textbf{Joint Accuracy (\%)}} &
\multicolumn{4}{c}{\textbf{Video-Level Breakdown (\%)}} \\
\cmidrule(lr){1-2}\cmidrule(lr){3-6}\cmidrule(lr){7-10}
Name & Think & $S\!+\!A\uparrow$ & $S\!+\!P\uparrow$ & $A\!+\!P\uparrow$ & $S\!+\!A\!+\!P\uparrow$ & 0/3 & 1/3 & 2/3 & 3/3 \\

\BreakdownBlockRow{Proprietary Models}
Seed1.8 & \cmark & \Best{74.88} & \Best{89.55} & \Best{72.58} & \Best{70.81} & 0.46 & 4.14 & 24.59 & \Best{70.81} \\
Qwen3.5-Plus & \cmark & \Second{71.23} & \Third{82.88} & \Second{71.56} & \Second{67.22} & 1.25 & 7.50 & 24.03 & \Second{67.22} \\
Seed1.8 & \xmark & 60.82 & \Second{89.15} & \Third{57.92} & \Third{56.94} & 0.66 & 5.33 & 37.08 & \Third{56.94} \\
Gemini2.5-Pro & \cmark & \Third{62.76} & 62.17 & 46.45 & 43.95 & 2.57 & 13.95 & 39.54 & 43.95 \\
Gemini2.5-Flash & \xmark & 52.99 & 64.89 & 48.39 & 44.05 & 5.39 & 16.44 & 34.12 & 44.05 \\
GPT5.4 & \xmark & 29.85 & 47.01 & 25.64 & 21.17 & 10.65 & 29.19 & 38.99 & 21.17 \\
Claude4.6-Sonnet & \xmark & 21.04 & 42.21 & 22.68 & 16.63 & 13.87 & 33.46 & 36.03 & 16.63 \\
Claude4.6-Opus & \cmark & 25.79 & 35.39 & 27.43 & 18.88 & 18.03 & 31.12 & 31.97 & 18.88 \\
Gemini3-Flash & \xmark & 27.74 & 31.89 & 21.76 & 15.91 & 15.91 & 34.52 & 33.66 & 15.91 \\
Seed1.6 & \xmark & 13.63 & 40.16 & 11.64 & 9.38 & 18.35 & 34.97 & 37.30 & 9.38 \\
Grok4.1 & \xmark & 5.19 & 5.06 & 5.26 & 0.72 & 42.47 & 43.46 & 13.35 & 0.72 \\
\addlinespace[2pt]

\BreakdownBlockRow{Open-source Models}
Qwen3.5-397B-A17B & \cmark & \Second{41.62} & \Best{62.26} & \Best{42.14} & \Best{35.77} & 5.46 & 20.05 & 38.72 & \Best{35.77} \\
Qwen3.5-35B-A3B & \cmark & \Best{42.46} & \Second{57.08} & \Second{38.89} & \Second{33.80} & 7.67 & 21.49 & 37.04 & \Second{33.80} \\
KimiK2.5 & \xmark & \Third{33.53} & \Third{49.77} & \Third{30.51} & \Third{25.51} & 10.06 & 27.15 & 37.28 & \Third{25.51} \\
Qwen3VL-32B-I & \xmark & 7.30 & 37.34 & 7.63 & 5.46 & 21.30 & 37.34 & 35.90 & 5.46 \\
InternVL3.5-38B & \xmark & 16.40 & 22.62 & 20.22 & 11.35 & 29.58 & 33.89 & 25.19 & 11.35 \\
Qwen3VL-8B-I & \xmark & 6.92 & 16.21 & 7.58 & 3.49 & 35.51 & 40.78 & 20.22 & 3.49 \\
InternVL3.5-8B & \xmark & 8.15 & 9.60 & 10.06 & 3.81 & 39.84 & 39.97 & 16.37 & 3.81 \\
InternVL3-38B & \xmark & 8.81 & 10.91 & 7.43 & 2.96 & 40.96 & 37.80 & 18.28 & 2.96 \\
KimiVL-A3B-T & \cmark & 6.96 & 6.67 & 6.96 & 2.17 & 37.83 & 45.94 & 14.06 & 2.17 \\
Qwen3VL-32B-T & \cmark & 5.92 & 6.58 & 6.71 & 1.91 & 41.25 & 43.36 & 13.49 & 1.91 \\
LLaVACriticR1-7B & \cmark & 6.99 & 5.79 & 6.85 & 1.73 & 41.78 & 42.05 & 14.44 & 1.73 \\
Qwen2.5VL-7B-I & \xmark & 6.47 & 8.03 & 7.42 & 2.11 & 42.07 & 40.23 & 15.59 & 2.11 \\
GLM4.6V-Flash & \xmark & 5.02 & 10.43 & 7.99 & 3.04 & 44.36 & 38.28 & 14.32 & 3.04 \\
Qwen3VL-4B-T & \cmark & 7.23 & 5.79 & 5.85 & 1.64 & 41.62 & 42.80 & 13.94 & 1.64 \\
Qwen3VL-8B-T & \cmark & 6.84 & 5.65 & 6.25 & 1.64 & 42.47 & 42.08 & 13.81 & 1.64 \\
ThinkLiteVL-7B & \cmark & 6.04 & 5.37 & 5.84 & 1.33 & 42.20 & 43.20 & 13.27 & 1.33 \\
MiMoVL-7B-RL & \xmark & 7.43 & 6.38 & 8.42 & 2.76 & 46.09 & 37.21 & 13.94 & 2.76 \\
Qwen3VL-4B-I & \xmark & 6.05 & 7.36 & 7.63 & 2.76 & 45.69 & 38.79 & 12.75 & 2.76 \\
KimiVL-A3B-I & \xmark & 8.25 & 5.98 & 6.87 & 2.75 & 47.77 & 36.63 & 12.85 & 2.75 \\
MiMoVL-7B-SFT & \xmark & 6.05 & 6.57 & 6.44 & 2.24 & 47.14 & 38.26 & 12.36 & 2.24 \\
Molmo2-8B & \xmark & 6.18 & 6.18 & 5.65 & 1.88 & 46.51 & 39.25 & 12.37 & 1.88 \\
\bottomrule
\end{tabularx}
\vspace{-1.5em}
\end{table*}

\begin{table}[ht]
\vspace{-0.3em}
\centering
\caption{Edge baseline, measured by the mean probability (\%) assigned to the ground-truth candidate.}
\label{tab:edge_baseline}
\vspace{-0.4em}
\footnotesize
\setlength{\tabcolsep}{4pt}
\renewcommand{\arraystretch}{1.08}
\begin{tabular}{@{}lrrrr@{}}
\toprule
\textbf{Setting} & \textbf{S} & \textbf{A} & \textbf{P} & \textbf{Mean} \\
\midrule
Edge-1 & 26.69 & 26.06 & 26.29 & 26.35 \\
Edge-4 & 25.91 & 24.91 & 25.66 & 25.49 \\
\bottomrule
\end{tabular}
\vspace{-1.0em}
\end{table}

\subsubsection{Single-Frame Baseline}
\label{sec:single_frame_baseline}

For Seed1.8 and Gemini2.5-Pro, we represent every beginning, ending, and candidate clip by its midpoint frame. Table~\ref{tab:single_frame_baseline} compares this setting with the original multi-frame evaluation.

\begin{table}[ht]
\vspace{-0.5em}
\centering
\caption{Original multi-frame evaluation versus the single-frame baseline. All values are percentages.}
\label{tab:single_frame_baseline}
\vspace{-0.4em}
\setlength{\tabcolsep}{2.4pt}
\renewcommand{\arraystretch}{1.08}
\resizebox{\linewidth}{!}{%
\begin{tabular}{@{}lrrrrrrr@{}}
\toprule
\textbf{Name} & \textbf{S$\uparrow$} & \textbf{A$\uparrow$} & \textbf{P$\uparrow$} & \textbf{Mean$\uparrow$} & \textbf{$\geq$1$\uparrow$} & \textbf{$\geq$2$\uparrow$} & \textbf{3/3$\uparrow$} \\
\midrule
Seed1.8 & 96.25 & 61.93 & 92.11 & 83.43 & 99.35 & 94.02 & 56.94 \\
\textbf{Seed1.8 (Single)} & \textbf{38.72} & \textbf{6.97} & \textbf{6.25} & \textbf{17.31} & \textbf{44.64} & \textbf{6.77} & \textbf{0.53} \\
\midrule
Gemini2.5-Pro & 90.33 & 66.67 & 67.78 & 74.92 & 97.44 & 83.49 & 43.95 \\
\textbf{Gemini2.5-Pro (Single)} & \textbf{40.50} & \textbf{8.81} & \textbf{7.89} & \textbf{19.07} & \textbf{48.13} & \textbf{8.55} & \textbf{0.53} \\
\bottomrule
\end{tabular}%
}
\vspace{-1.0em}
\end{table}

\section{Human Baseline}
\label{sec:human_baseline}

\begin{figure*}[htb]
  \centering
  \includegraphics[width=0.98\textwidth]{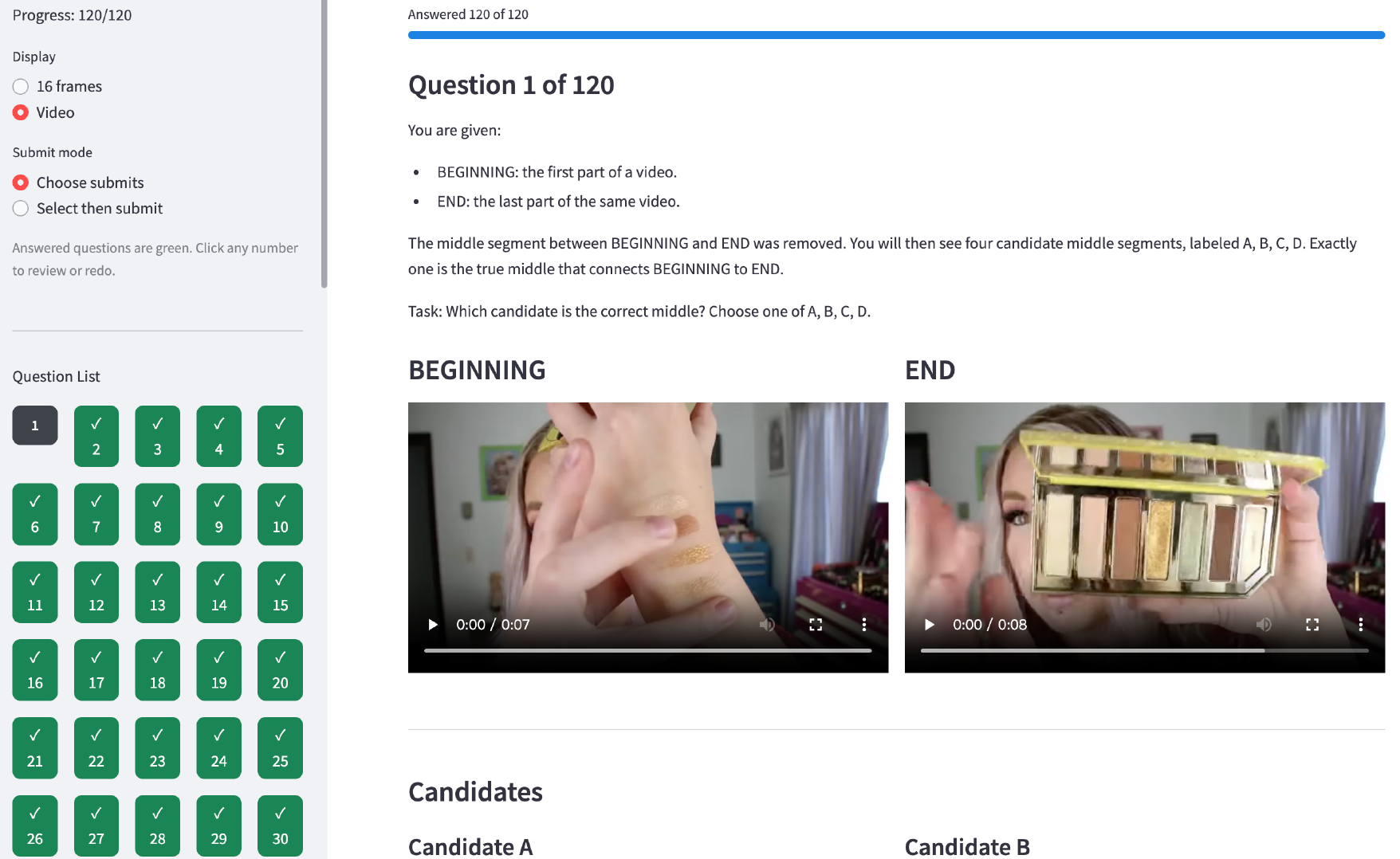}
  \caption{Human annotation interface. Annotators viewed the same clips as the models.}
  \label{fig:annotation_interface}
\end{figure*}

We estimate a Human Baseline on 100 randomly selected videos from \bench. Five annotators completed the task after a short briefing. A Streamlit interface showed the same inputs as model evaluation: beginning clip, ending clip, and four candidate middle clips (Figure~\ref{fig:annotation_interface}).

Each annotator answered 120 questions. Every question received two independent judgments. We count a question as correct only if both annotators selected the ground-truth clip; all disagreements count as incorrect.

Under this strict rule, humans solve 96/100 Semantic, 98/100 Alignment, and 97/100 Progression questions, for 291/300 = 97.0\% mean dimension accuracy. At the video level, humans solve at least one dimension for 100/100 videos, at least two for 99/100, and all three for 92/100.

\begin{table*}[tb]
\centering
\caption{Source-level filtering statistics.}
\vspace{-0.5em}
\label{tab:filtering_pipeline}
\footnotesize
\setlength{\tabcolsep}{3.8pt}
\renewcommand{\arraystretch}{1.08}
\begin{tabular}{@{}l r *{9}{c}@{}}
\toprule
\multirow{2}{*}{\textbf{Source}} & \multirow{2}{*}{\textbf{Input}} &
\multicolumn{2}{c}{\textbf{Duration}} &
\multicolumn{2}{c}{\textbf{LLM}} &
\multicolumn{2}{c}{\textbf{Quality}} &
\multicolumn{2}{c}{\textbf{Motion}} &
\multirow{2}{*}{\textbf{Filtered (\%)}} \\
\cmidrule(lr){3-4}\cmidrule(lr){5-6}\cmidrule(lr){7-8}\cmidrule(lr){9-10}
& & \textbf{Count} & \textbf{Share (\%)} & \textbf{Count} & \textbf{Share (\%)} & \textbf{Count} & \textbf{Share (\%)} & \textbf{Count} & \textbf{Share (\%)} & \\
\midrule
CaReBench & 1,000 & 437 & 43.70 & 73 & 12.97 & 86 & 17.55 & 233 & 57.67 & 90.60 \\
Daily-Omni & 653 & 0 & 0.00 & 222 & 34.00 & 154 & 35.73 & 161 & 58.12 & 93.42 \\
EgoLife & 2,347 & 1 & 0.04 & 1,391 & 59.29 & 0 & 0.00 & 265 & 27.87 & 81.38 \\
FAVOR & 1,776 & 503 & 28.32 & 503 & 39.51 & 104 & 13.63 & 414 & 62.82 & 91.84 \\
LVD-2M & 82,196 & 75,252 & 91.55 & 3,843 & 55.34 & 156 & 5.03 & 192 & 6.52 & 99.37 \\
MiraData & 1,972 & -- & -- & 705 & 35.75 & 31 & 2.49 & 4 & 0.33 & 89.96 \\
Video-TT & 791 & -- & -- & 265 & 33.50 & 151 & 35.53 & 144 & 52.55 & 88.75 \\
\bottomrule
\end{tabular}
\vspace{-0.5em}
\end{table*}

\section{Source Datasets}
\label{sec:filtering_pipeline}

\subsection{Dataset Overview}
\label{sec:source_dataset_overview}

\paragraph{LVD-2M~\citep{xiong2024lvd2m}}
is a large-scale open-domain video dataset collected from the web. It contains approximately 2 million long-take videos, each lasting more than 10 seconds with continuous motion. The dataset provides temporally dense captions generated through a hierarchical annotation pipeline, describing events at fine-grained intervals. Videos cover varied scenes, camera motion, and durations, supporting models that learn extended spatiotemporal dependencies.

\paragraph{Video-TT~\citep{zhang2025videott}}
contains 1,000 short video clips sourced from YouTube Shorts. Each video is paired with one open-ended question and four natural adversarial variants for visual and narrative comprehension. The dataset spans daily life, comedy, sports, and 18 question types covering element attributes, event sequencing, and plot understanding. Videos were filtered for length (under 65 seconds), content safety, and frame sampling quality, with rationales curated to support reasoning under question variation.

\paragraph{FAVOR-Bench~\citep{tu2025favorbench}}
focuses on fine-grained actions and temporally sensitive visual states. It contains short to medium-length clips that capture changes in object states or human motions. Videos are annotated for temporal dynamics, scene transitions, and event variations, supporting analysis of temporal reasoning.

\paragraph{CaReBench~\citep{xu2026carebench}}
is a curated video dataset with temporal descriptions. It consists of moderate-length videos depicting indoor and outdoor everyday activities. Each video includes annotations for scene content and event transitions, with structured temporal information separating spatial and temporal aspects of the scene. The dataset supports fine-grained evaluation of video comprehension and retrieval.

\paragraph{Daily-Omni~\citep{zhou2025dailyomni}}
consists of daily-life videos paired with captions describing human activities and object interactions. Videos are recorded across different environments and scenarios, capturing common daily behaviors. The annotations are aligned to observable actions and interactions.

\paragraph{EgoLife~\citep{yang2025egolife}}
is a dataset of first-person life-logging videos recorded from head-mounted or body-mounted cameras. It contains continuous footage of everyday activities across multiple participants, capturing egocentric perspectives, motion patterns, and interactions. Audio is included in the original recordings, but the dataset focuses mainly on visually grounded action annotations.

\paragraph{MiraData~\citep{ju2024miradata}}
provides densely captioned videos with event descriptions. Clips vary in length and are filtered and deduplicated for quality and continuity. Structured captions describe objects, actions, background, and motion patterns at multiple granularities. The dataset contains hundreds of thousands of videos, making it suitable for temporal reasoning and continuous video understanding.

\subsection{Filtering Statistics}
\label{sec:filtering_statistics}

Table~\ref{tab:filtering_pipeline} reports the filtering ratio at each stage. For sources without captions, we first generate fine-grained captions with Gemini-2.5-Pro, then use the available or generated captions for LLM suitability screening. Overall, duration filtering removes the largest number of videos, mainly because long web-scale sources contain many clips outside our target length range. Among later stages, LLM suitability and motion filtering are the most selective.

Retained videos then enter the same gap-generation stage: sample a central missing span, validate it with Farneb\"ack optical flow, and allow up to three resampling attempts. Stage percentages are filter-conditional; the total column is relative to the original source input.

\subsection{Human Validation}
\label{sec:filtered_video_human_validation}

\begin{figure*}[htb]
  \centering
  \includegraphics[width=0.98\textwidth]{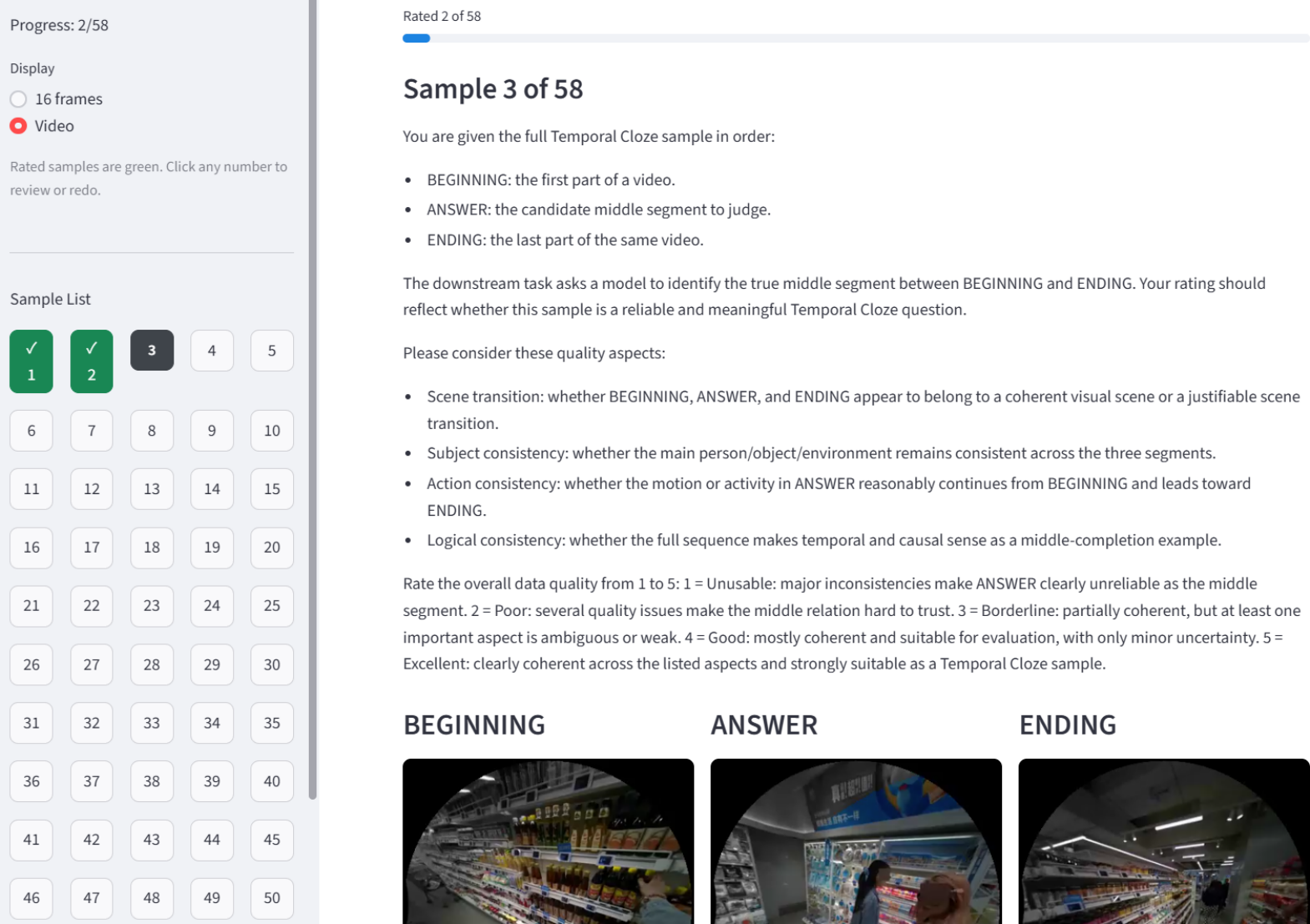}
  \caption{Streamlit interface for filtered-video quality review.}
  \label{fig:annotation_interface_quality}
\end{figure*}

We conduct a small-scale human validation on the filtered videos. Five annotators each review 60 videos, for 300 videos in total. Each annotator rates whether a video is suitable for the cloze task on a five-point scale: 1 = Unusable, 2 = Somewhat unsuitable, 3 = Acceptable, 4 = Somewhat suitable, and 5 = Very suitable. The Streamlit front end used for this review is shown in Figure~\ref{fig:annotation_interface_quality}.

\begin{table*}[bth]
\centering
\caption{Gemini2.5-Pro statistics by source.}
\vspace{-0.5em}
\label{tab:gemini_source_stats}
\footnotesize
\renewcommand{\arraystretch}{0.98}
\setlength{\tabcolsep}{4.2pt}
\begin{tabularx}{\textwidth}{@{}l *{7}{>{\centering\arraybackslash}X}@{}}
\toprule
\multirow{2}{*}{\textbf{Source}} &
\multicolumn{4}{c}{\textbf{Dimension Accuracy (\%)}} &
\multicolumn{3}{c}{\textbf{Cumulative Accuracy (\%)}} \\
\cmidrule(lr){2-5}\cmidrule(l){6-8}
& \textbf{S$\uparrow$} & \textbf{A$\uparrow$} & \textbf{P$\uparrow$} & \textbf{Mean$\uparrow$} & \textbf{$\geq\!1\uparrow$} & \textbf{$\geq\!2\uparrow$} & \textbf{3/3$\uparrow$} \\
\midrule
CaReBench & 91.49 & 64.89 & 62.77 & 73.05 & 94.68 & 81.91 & 42.55 \\
Daily-Omni & 93.02 & 79.07 & 60.47 & 77.52 & 97.67 & 83.72 & 51.16 \\
EgoLife & 83.52 & 59.73 & 66.13 & 69.79 & 95.88 & 78.03 & 35.47 \\
FAVOR-Bench & 95.17 & 64.83 & 75.86 & 78.62 & 99.31 & 84.83 & 51.72 \\
LVD-2M & 93.19 & 68.54 & 69.71 & 77.15 & 98.44 & 87.35 & 45.91 \\
MiraData & 94.95 & 78.28 & 69.70 & 80.98 & 98.99 & 90.40 & 53.54 \\
Video-TT & 86.52 & 62.92 & 56.18 & 68.54 & 95.51 & 71.91 & 38.20 \\
\midrule
\textbf{Overall} & \textbf{90.33} & \textbf{66.67} & \textbf{67.78} & \textbf{74.92} & \textbf{97.43} & \textbf{83.49} & \textbf{43.95} \\
\bottomrule
\end{tabularx}
\vspace{-1em}
\end{table*}

\begin{table*}[tb]
\centering
\caption{Qwen3.5-35B-A3B statistics by source.}
\vspace{-0.5em}
\label{tab:qwen_source_stats}
\footnotesize
\renewcommand{\arraystretch}{0.98}
\setlength{\tabcolsep}{4.2pt}
\begin{tabularx}{\textwidth}{@{}l *{7}{>{\centering\arraybackslash}X}@{}}
\toprule
\multirow{2}{*}{\textbf{Source}} &
\multicolumn{4}{c}{\textbf{Dimension Accuracy (\%)}} &
\multicolumn{3}{c}{\textbf{Cumulative Accuracy (\%)}} \\
\cmidrule(lr){2-5}\cmidrule(l){6-8}
& \textbf{S$\uparrow$} & \textbf{A$\uparrow$} & \textbf{P$\uparrow$} & \textbf{Mean$\uparrow$} & \textbf{$\geq\!1\uparrow$} & \textbf{$\geq\!2\uparrow$} & \textbf{3/3$\uparrow$} \\
\midrule
CaReBench & 61.70 & 33.33 & 51.06 & 48.70 & 80.65 & 50.54 & 15.05 \\
Daily-Omni & 86.05 & 62.79 & 81.40 & 76.75 & 95.35 & 83.72 & 51.16 \\
EgoLife & 79.86 & 53.09 & 71.10 & 68.02 & 94.95 & 73.62 & 35.55 \\
FAVOR-Bench & 76.92 & 51.75 & 75.17 & 67.95 & 93.66 & 75.35 & 36.62 \\
LVD-2M & 70.23 & 47.96 & 65.24 & 61.14 & 89.88 & 64.40 & 29.38 \\
MiraData & 76.26 & 61.42 & 74.49 & 70.72 & 94.36 & 78.97 & 39.49 \\
Video-TT & 83.15 & 56.18 & 87.64 & 75.66 & 97.75 & 84.27 & 44.94 \\
\midrule
\textbf{Overall} & \textbf{75.10} & \textbf{51.55} & \textbf{69.96} & \textbf{65.54} & \textbf{92.33} & \textbf{70.83} & \textbf{33.80} \\
\bottomrule
\end{tabularx}
\vspace{-1em}
\end{table*}

\begin{table}[hbt]
\vspace{-0.5em}
\centering
\caption{Human validation of filtered video suitability.}
\vspace{-0.8em}
\label{tab:human_quality_validation}
\footnotesize
\setlength{\tabcolsep}{2.4pt}
\renewcommand{\arraystretch}{1.08}
\begin{tabularx}{\linewidth}{@{}l@{\hspace{1.2pt}}*{7}{>{\centering\arraybackslash}X}@{}}
\toprule
\textbf{Annotator} & \textbf{1} & \textbf{2} & \textbf{3} & \textbf{4} & \textbf{5} & \textbf{Total} & \textbf{Avg.} \\
\midrule
No.1 & 0 & 2 & 10 & 14 & 34 & 60 & 4.33 \\
No.2 & 0 & 5 & 17 & 27 & 11 & 60 & 3.73 \\
No.3 & 0 & 2 & 3  & 34 & 21 & 60 & 4.23 \\
No.4 & 0 & 4 & 4  & 12 & 40 & 60 & 4.47 \\
No.5 & 0 & 3 & 16 & 25 & 16 & 60 & 3.90 \\
\midrule
\rowcolor{gray!10}
\textbf{Overall} & \textbf{0} & \textbf{16} & \textbf{50} & \textbf{112} & \textbf{122} & \textbf{300} & \textbf{4.13} \\
\rowcolor{gray!5}
\textbf{Share (\%)} & 0.0 & 5.3 & 16.7 & 37.3 & 40.7 & 100.0 & -- \\
\bottomrule
\end{tabularx}
\vspace{-1.5em}
\end{table}

Table~\ref{tab:human_quality_validation} shows that the filtered videos are generally suitable for the cloze task. The overall mean rating is 4.13 out of 5, with 234/300 ratings (78.0\%) in the two positive categories and 284/300 ratings (94.7\%) at least acceptable. Importantly, rating 1 (Unusable) occurs 0 times, indicating that the filtering pipeline removes clearly invalid cases.

\section{Statistics by Sources}
\label{sec:source_statistics}

Tables~\ref{tab:gemini_source_stats} and~\ref{tab:qwen_source_stats} report source-level results for two strong models. Metrics follow Table~\ref{tab:main_results}: S/A/P are dimension accuracies, Mean averages them, and cumulative columns count dimensions solved on the same video.

Gemini2.5-Pro is consistently strong on Semantic, while Alignment remains lower. MiraData gives its best 3/3 score.

Qwen3.5-35B-A3B varies more by source and is strongest on Daily-Omni and Video-TT. For both models, high $\geq\!1$ accuracy with much lower 3/3 accuracy shows that failures are often partial.

\begin{figure*}[t]
  \centering
  \includegraphics[width=\textwidth]{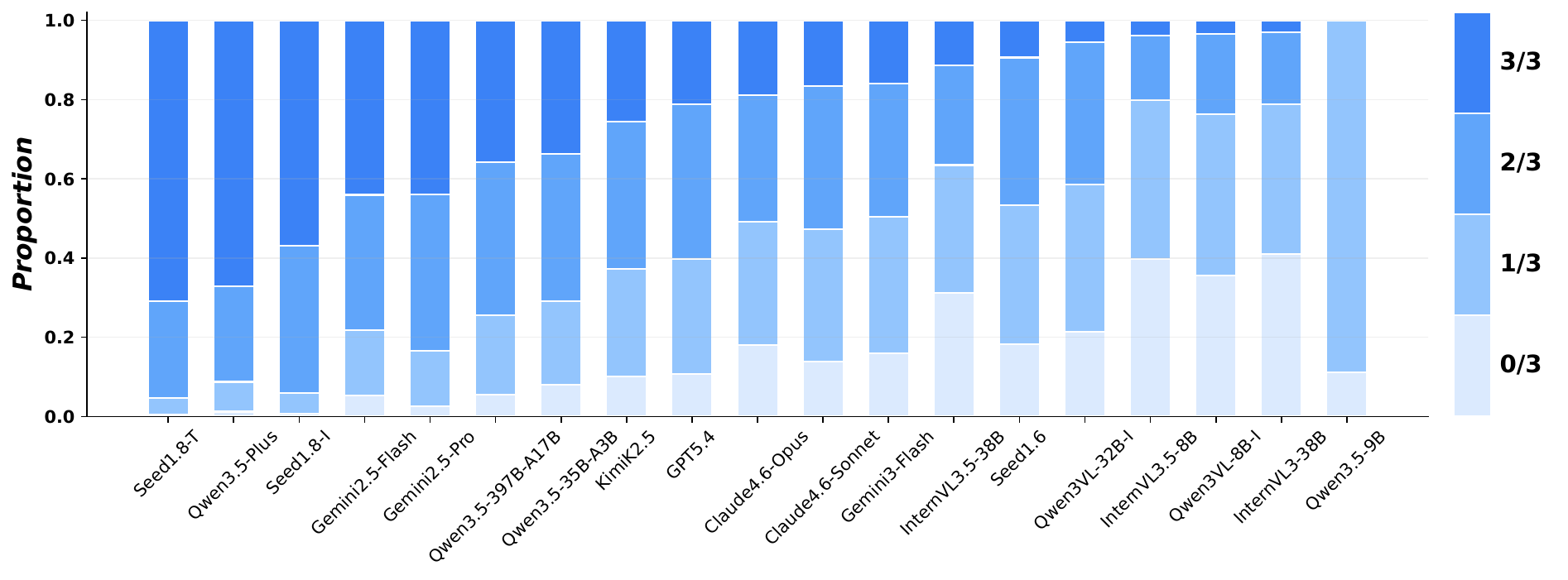}
  \vspace{-1.8em}
  \caption{Video-level breakdown across models.}
  \label{fig:breakdown}
  \vspace{-1em}
\end{figure*}

\section{Additional Analysis on Main Results}
\label{sec:video_level_metrics}

\subsection{Joint Accuracy}
\label{sec:joint_accuracy}

Pairs involving Alignment are consistently weaker than $S{+}P$. Proprietary models average 40.54\% on $S{+}A$ and 37.39\% on $A{+}P$, versus 53.67\% on $S{+}P$; open-source models show the same pattern at lower levels (11.93\%, 12.06\%, and 16.79\%). Seed~1.8 (thinking) leads proprietary joint accuracy, while Qwen3.5-35B-A3B and Qwen3.5-397B-A17B lead open-source models.

\subsection{Video-Level Breakdown}
\label{sec:video_level_breakdown}

The proprietary--open-source gap is larger at the video level: 1.9$\times$ in mean dimension accuracy (62.19\% vs.\ 32.51\%) but 4.6$\times$ in 3/3 accuracy (33.24\% vs.\ 7.15\%). Proprietary failures are often narrow 2/3 misses (31.88\% vs.\ 19.34\%); open-source failures concentrate in 0/3 and 1/3 cases. Qwen3.5-397B-A17B reaches 38.72\% in 2/3 but 35.77\% in 3/3, while the best proprietary models reach 67.22\%--70.81\% in 3/3. For Seed~1.8, thinking mainly shifts videos from 2/3 to 3/3 (37.08\% to 24.59\%, 56.94\% to 70.81\%).

\subsection{Effect of Thinking}
\label{sec:thinking_effect}

\begin{figure*}[tbh]
\centering
\begin{tabular}{@{}c@{\hspace{0.025\textwidth}}c@{\hspace{0.025\textwidth}}c@{}}
\includegraphics[width=0.31\textwidth]{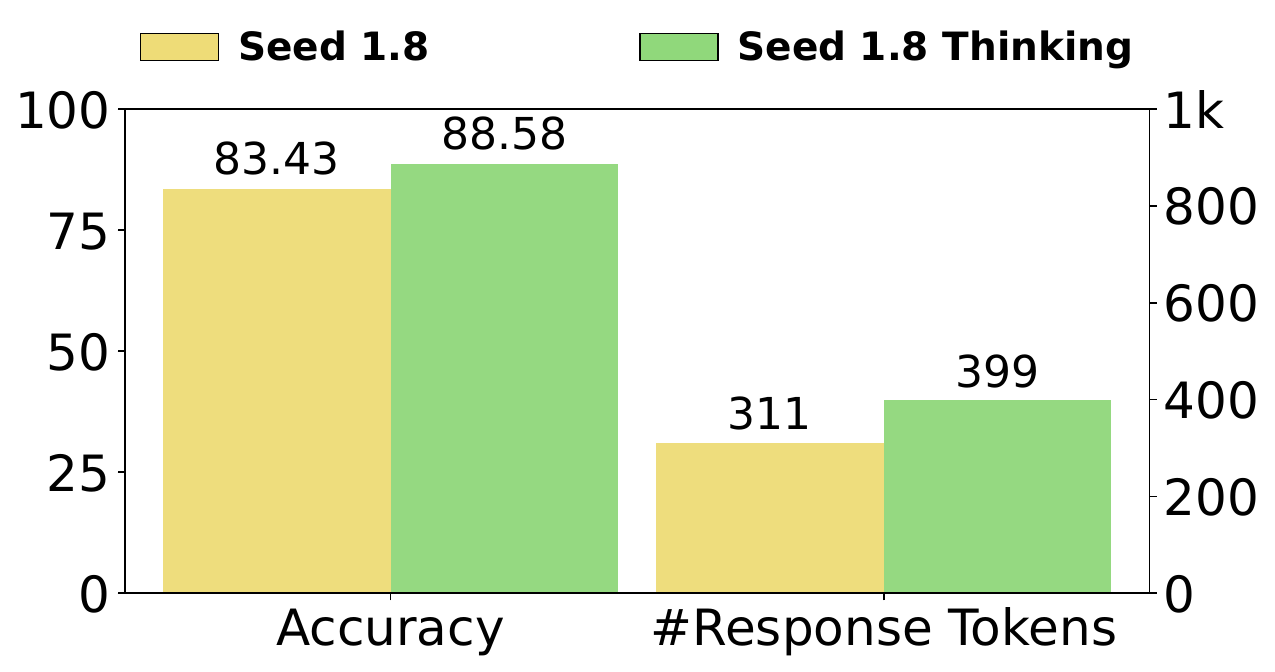} &
\includegraphics[width=0.31\textwidth]{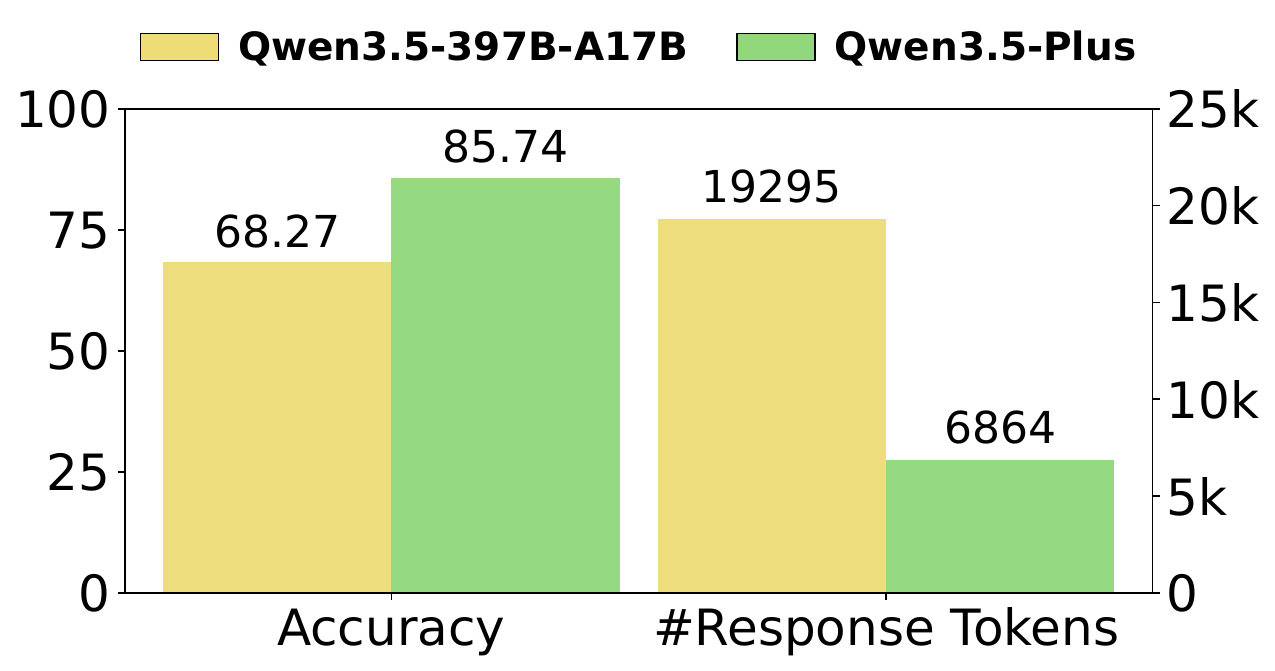} &
\includegraphics[width=0.31\textwidth]{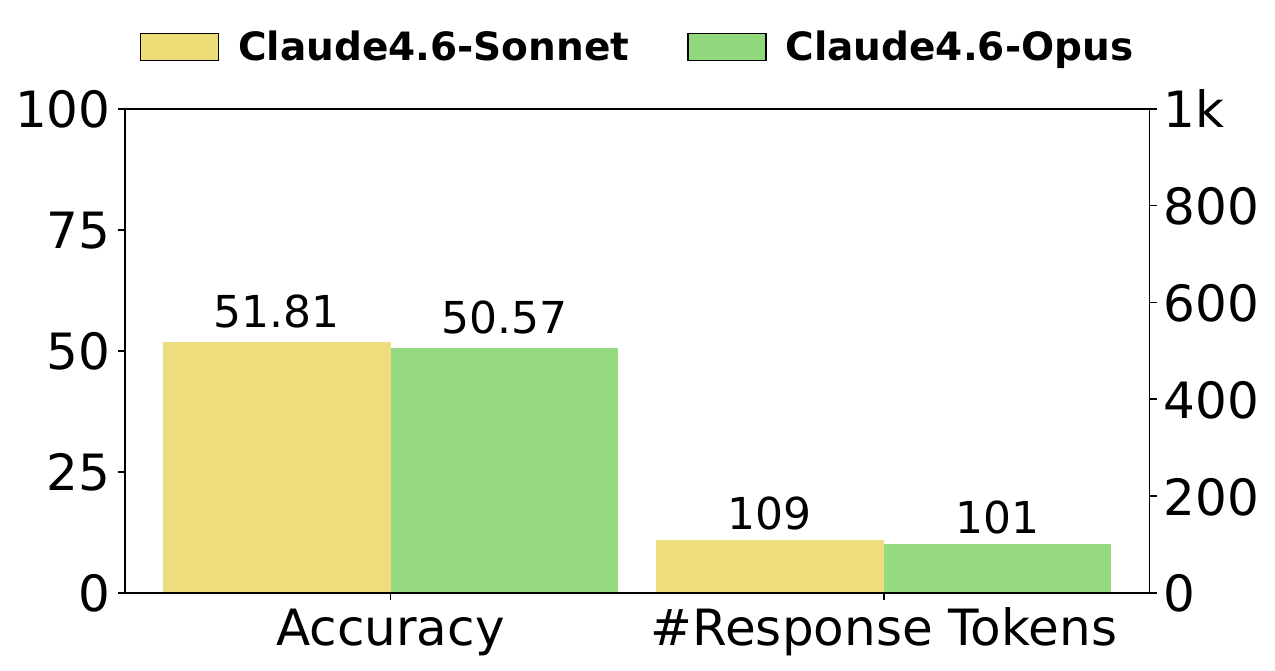} \\
{\small Seed with and without thinking} &
{\small Qwen open-weight and hosted} &
{\small Claude Sonnet \& Opus}
\end{tabular}
\vspace{-0.6em}
\caption{Thinking and mode comparisons: mean accuracy vs.\ average response tokens.}
\label{fig:thinking_mode}
\end{figure*}

The Think column in Table~\ref{tab:main_results} mixes proprietary reasoning modes and open-source thinking checkpoints; we use matched pairs where possible. Table~\ref{tab:thinking_effect} reports percentage-point changes from non-thinking or instruct baselines, and Figure~\ref{fig:thinking_mode} compares representative accuracy--token tradeoffs.

\begin{table}[tb]
\vspace{-0.5em}
\centering
\caption{Effect of thinking variants.}
\label{tab:thinking_effect}
\vspace{-0.5em}
\footnotesize
\setlength{\tabcolsep}{3.2pt}
\renewcommand{\arraystretch}{1.08}
\providecommand{\Gain}[1]{\textcolor{green!38!black}{+#1}}
\providecommand{\Drop}[1]{\textcolor{red!65!black}{-#1}}
\resizebox{0.98\linewidth}{!}{%
\begin{tabular}{@{}lrrrr@{}}
\toprule
\textbf{Pair} & \textbf{$\Delta$S} & \textbf{$\Delta$A} & \textbf{$\Delta$P} & \textbf{$\Delta$Mean} \\
\midrule
Seed1.8 & \Drop{1.18} & \Gain{14.99} & \Gain{1.64} & \Gain{5.15} \\
Qwen3VL-32B & \Drop{26.30} & \Gain{14.47} & \Drop{37.53} & \Drop{16.46} \\
Qwen3VL-8B & \Drop{6.49} & \Gain{10.32} & \Drop{20.87} & \Drop{5.68} \\
Qwen3VL-4B & \Gain{7.43} & \Gain{5.13} & \Drop{9.53} & \Gain{1.01} \\
KimiVL-A3B & \Gain{9.08} & \Drop{1.26} & \Gain{1.85} & \Gain{3.24} \\
\bottomrule
\end{tabular}%
}
\vspace{-1.5em}
\end{table}

Thinking helps most when a strong base model stays visually grounded. Seed~1.8 gains 5.15 mean-accuracy points and 13.87 full-joint points, mainly from a 14.99-point Alignment gain; Semantic changes little.

Open-source pairs are mixed. Qwen3VL thinking checkpoints improve Alignment at all matched sizes but often lose more on Semantic and Progression. Qwen3VL-32B-T, for example, gains 14.47 Alignment points but drops 26.30 on Semantic and 37.53 on Progression relative to Qwen3VL-32B-I.

KimiVL-A3B is milder: thinking improves mean accuracy by 3.24 points, mainly through Semantic, but slightly lowers full joint accuracy.
Overall, thinking is not uniformly beneficial on \bench; it helps only when extra reasoning improves temporal timing without losing visual or progression constraints.

\section{Prompts}
\label{sec:temporal-cloze-eval-prompt}

\subsection{Temporal Cloze Suitability Prompt}
\lstset{
  basicstyle=\ttfamily\footnotesize,
  breaklines=true,
  columns=fullflexible,
  keepspaces=true,
  frame=single,
  rulecolor=\color{blue!75!black},
  backgroundcolor=\color{blue!5!white},
  framesep=5pt
}
\begin{lstlisting}
O3_PROMPT = """[Temporal Cloze Task Suitability]

You are judging whether the video described by the caption is suitable for a temporal cloze benchmark. In this benchmark, a video is split into three ordered visual segments:
- BEGINNING: the context shown before the missing middle.
- ANSWER: the missing middle segment that a model should choose or predict.
- ENDING: the context shown after the missing middle.

A good example lets a viewer infer a plausible ANSWER from BEGINNING and ENDING through visual temporal and causal continuity. Judge only what can be supported by visible actions or events in the caption. Ignore audio-only information, speech or dialogue content, subtitles, narration, background music, creator commentary, and any purely emotional or aesthetic judgment unless it is visually grounded.

Please consider these quality aspects:
- Scene transition: whether BEGINNING, ANSWER, and ENDING appear to belong to a coherent visual scene or a justifiable scene transition.
- Subject consistency: whether the main person, object, or environment remains consistent across the three segments.
- Action consistency: whether the motion or activity in ANSWER reasonably continues from BEGINNING and leads toward ENDING.
- Logical consistency: whether the full sequence makes temporal and causal sense as a middle-completion example.

PASS only if:
- The caption describes a concrete visual process, action, interaction, transformation, or event with ordered steps.
- The likely ANSWER segment would be visually distinguishable and meaningfully constrained by BEGINNING and ENDING.
- BEGINNING, ANSWER, and ENDING would form one coherent temporal sequence with stable subjects and a reasonable causal or physical progression.

REJECT if any of these apply:
- The caption is mostly static description, scenery, appearance, atmosphere, or a montage without a clear ordered visual action.
- The event depends mainly on dialogue, audio, text, subtitles, narration, or hidden intent rather than visible motion.
- The middle could be almost anything, is too ambiguous, or cannot be inferred from before/after visual context.
- There are abrupt unrelated scene cuts, inconsistent subjects or environments, or disconnected actions that would make BEGINNING, ANSWER, and ENDING incoherent.
- The video only shows repeated motion with no meaningful progress, or a single state with little temporal change.

Caption:
{caption}

Return valid JSON only, using one of these shapes:
{{"pass": true, "reason": "one or two concise sentences naming the decisive quality aspects"}}
{{"pass": false, "reason": "one or two concise sentences naming the decisive quality aspects"}}
"""
\end{lstlisting}

\subsection{Temporal Cloze Evaluation Prompt}
\lstset{
  basicstyle=\ttfamily\footnotesize,
  breaklines=true,
  columns=fullflexible,
  keepspaces=true,
  frame=single,
  rulecolor=\color{blue!75!black},
  backgroundcolor=\color{blue!5!white},
  framesep=5pt
}
\begin{lstlisting}
eval_prompt = """
You are given:
- **BEGINNING**: the first part of a video (frames in temporal order).
- **END**: the last part of the same video (frames in temporal order).
The middle segment between BEGINNING 
and END was removed.

You will then see four candidate middle 
segments, labeled A, B, C, D.
Each candidate is a short clip;
exactly one is the true middle that 
connects BEGINNING to END.
The others are wrong.

Task: Which candidate is the correct middle? Choose one of A, B, C, D.

**Output JSON only** : {"answer": "<A, B, C, or D>", "reason": "<one or two sentences>"}
"""
\end{lstlisting}

\newpage

\end{document}